%% file: main.tex
\documentclass[runningheads]{llncs}
\input{preamble}

\begin{document}

\raggedbottom

\title{Reflection-Robust 6DoF Object Tracking with Light Fields}

\author{Nikolai Goncharov\orcidlink{0009-0004-3477-9638} \and
Donald G. Dansereau\orcidlink{0000-0003-2540-1639}}

\authorrunning{N.~Goncharov and D.~G.~Dansereau}

\institute{Australian Centre for Robotics, School of Aerospace, Mechanical and
  Mechatronic Engineering, University of Sydney, Sydney, NSW, Australia\\
\email{\{nikolai.goncharov, donald.dansereau\}@sydney.edu.au}}

\maketitle

\begin{abstract}
  Tracking the \gls{dof} pose of a moving rigid object is
  fundamental to robotics and autonomous driving, but existing trackers
  assume that object appearance is stable across a sequence, an assumption
  that breaks down on reflective surfaces whose appearance changes as they
  mirror the environment. We introduce a light field based
  reflection-robust \gls{dof} tracker that turns this apparent nuisance into a
  pose cue. Per frame, our method recovers depth robustly against
  reflections, back-projects it into a point cloud, and estimates
  surface normals. It then decomposes the object's
  view-dependent appearance into a diffuse albedo and the environment map it
  reflects, resulting in a relightable surface light field. Starting
  from a coarse initialization, we relight it by the recovered
  environment map and optimize the pose on the photometric loss.
  Because a moving object mirrors new parts of the scene, the
  environment map fills in as the sequence proceeds, sharpening this
  signal over time. To evaluate this approach, we introduce a light
  field tracking dataset re-rendered from a robotic manipulation
  benchmark at four controlled reflectivity levels, each paired with
  simulated depth that reproduces how consumer RGB-D sensors fail on
  shiny surfaces. Additionally, we evaluate on two captured light
  field sequences. Our method trails the strongest baselines on
  diffuse objects and is the only one that holds its accuracy on fully
  reflective objects, where every baseline degrades.

  \smallskip

  Code and data available at:
  \href{https://nagonch.github.io/lightfield_tracking/}{\texttt{nagonch.github.io/lightfield\_tracking}}.

\end{abstract}

\glsresetall

\input{sections/introduction}
\input{sections/related_work}
\input{sections/method}
\input{sections/dataset}
\input{sections/experiments}
\input{sections/conclusion}
\makeatletter\ifaccv@review\else\input{sections/acknowledgements}\fi\makeatother

\bibliographystyle{splncs04}
\bibliography{main}

\clearpage
\begingroup
\setcounter{figure}{0}
\setcounter{table}{0}
\renewcommand{\thetable}{S\arabic{table}}
\renewcommand{\thefigure}{S\arabic{figure}}
\renewcommand{\theHtable}{S\arabic{table}}
\renewcommand{\theHfigure}{S\arabic{figure}}
\phantomsection
\pdfbookmark[1]{Supplementary Material}{supp}
\begin{center}
  {\Large\bfseries Supplementary Material\par}
\end{center}
\input{supplementary}

\endgroup

\end{document}

%% file: preamble.tex
\usepackage{accv}

\usepackage{accvabbrv}

\usepackage{graphicx}
\usepackage{booktabs}

\usepackage[accsupp]{axessibility}  %

\usepackage[breaklinks,colorlinks,citecolor=accvblue]{hyperref}

\usepackage{orcidlink}
\usepackage{placeins}

\usepackage[acronym,nomain,nowarn]{glossaries}
\setacronymstyle{long-short}
\glsdisablehyper
\newacronym{dof}{6DoF}{six degrees of freedom}
\newacronym{slam}{SLAM}{simultaneous localization and mapping}
\newacronym{icp}{ICP}{iterative closest point}
\newacronym{pnp}{PnP}{perspective-$n$-point}
\newacronym{ransac}{RANSAC}{random sample consensus}
\newacronym{sift}{SIFT}{scale-invariant feature transform}
\newacronym{sdf}{SDF}{signed distance function}
\newacronym{sgbm}{SGBM}{semi-global block matching}
\newacronym{epi}{EPI}{epipolar plane image}
\newacronym{pca}{PCA}{principal component analysis}
\newacronym{brdf}{BRDF}{bidirectional reflectance distribution function}
\newacronym{add}{ADD}{average distance of model points}
\newacronym{auc}{AUC}{area under the accuracy-threshold curve}
\newacronym{ate}{ATE}{absolute translation error}
\newacronym{are}{ARE}{absolute rotation error}

\usepackage{mathtools}
\usepackage{multirow}
\usepackage{adjustbox}
\usepackage{pifont}
\usepackage{rotating}
\newcommand{\cmark}{\ding{51}}
\newcommand{\xmark}{\ding{55}}
\newcommand{\variable}[1]{\ensuremath{\boldsymbol{#1}}}
\newcommand{\coordframe}[2]{\prescript{\mathrm{#1}}{}{\variable{T}}_{\mathrm{#2}}}

\AtEndPreamble{
  \crefname{figure}{Fig.}{Figs.}
  \Crefname{figure}{Figure}{Figures}
  \crefname{equation}{Eq.}{Eqs.}
  \Crefname{equation}{Equation}{Equations}
}

\renewcommand{\topfraction}{0.92}
\renewcommand{\bottomfraction}{0.9}
\renewcommand{\textfraction}{0.06}
\renewcommand{\floatpagefraction}{0.85}

%% file: sections/introduction.tex
\section{Introduction}
\label{sec:intro}
Estimating the \gls{dof} pose of a moving rigid object is fundamental to
autonomous driving, dynamic \gls{slam}, and aerial robotics. On
reflective surfaces, however, the appearance of each pixel couples
pose with the object's geometry, material, and the surrounding
illumination. Many combinations of these unknowns explain the same
observation, and pose cannot be recovered from appearance alone
without modeling the illumination.

\input{figures/toy_reflection_motion_float}

Model-based and model-free \gls{dof} trackers~\cite{wen2024foundationpose,
labbe2022megapose, lin2024sam, wen2023bundlesdf, goncharov2025light}
assume stable object appearance and trustworthy depth, both of which
break down on specular surfaces.
Relightable reconstruction methods~\cite{srinivasan2021nerv,
liu2024bigs, gao2024relightable} recover appearance under novel
illumination but only offline, over many views. Work on the dynamics of
reflective surfaces~\cite{adato2010shape, chandraker2013object}
operates under assumptions that do not hold in unconstrained settings.
Reflections are generally treated as interference to be suppressed,
although they encode the object's geometry, motion, and
surroundings, and no prior tracker turns them into an online cue for
pose.

We show that the view-dependent reflection can serve as a pose cue
for object tracking (\cref{fig:reflection_as_pose_signal}). Our
method reconstructs and relights the tracked object online from a
single light field capture per frame. The tracking problem thus reduces to
aligning a correctly illuminated model. We make the following
contributions:

\begin{itemize}
  \item A \gls{dof} object tracking method that recovers pose while
    reconstructing the surrounding illumination and relighting the
    object. It stays accurate on reflective surfaces, for which
    the tracking quality of the state-of-the-art RGB-D trackers drops.

  \item A relightable object model of geometry, surface orientation,
    and material appearance, recovered from a single light field
    capture, whose appearance changes correctly as the object moves.

  \item An online decomposition of appearance into the object's
    diffuse color and its reflected environment, which gets updated
    as the object moves.

  \item A light field tracking dataset built by re-rendering a robotic
    manipulation benchmark~\cite{wen2020se} at several reflectivity
    levels with geometry, lighting, and pose identical across levels.
    We pair each frame with simulated depth that reproduces
    the quality degradation of
    consumer RGB-D sensors on reflective surfaces. Additionally, we provide two
    captured light field motion sequences of a diffuse and a
    reflective version of the same object (mustard bottle).
\end{itemize}

%% file: figures/toy_reflection_motion_float.tex
\begin{figure}[t]
  \centering
  \includegraphics[width=0.76\linewidth]{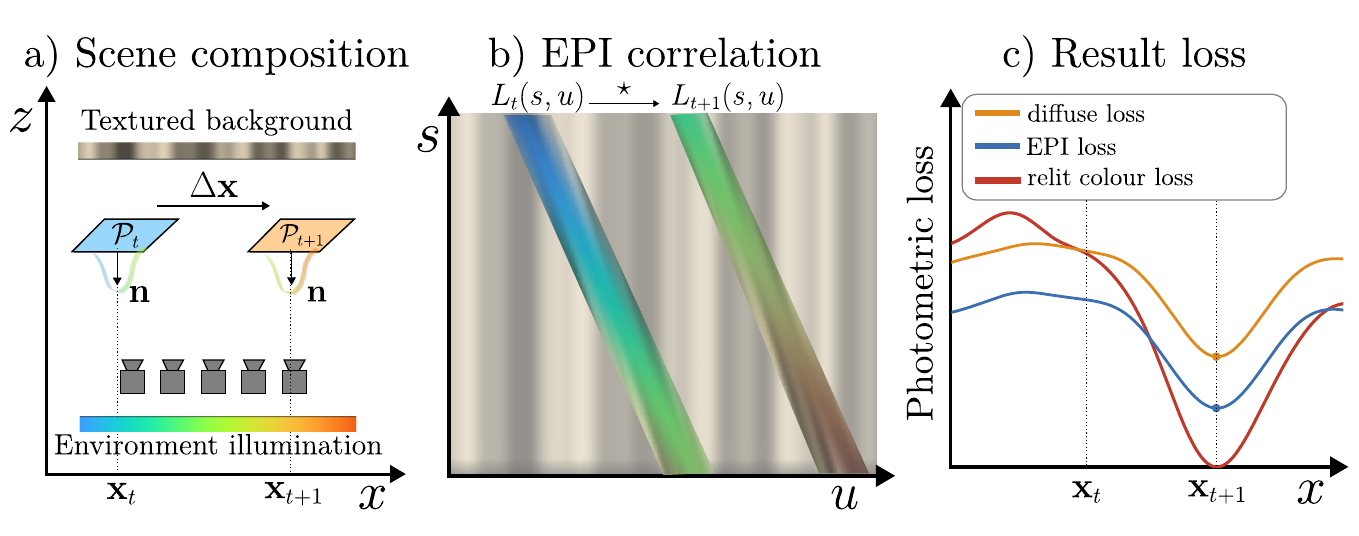}
  \caption[Reflection as a pose signal]{\textbf{Reflection as a
    pose signal.} \textbf{(a)} A reflective surface patch
    $\mathcal{P}$ in flatland, with normal $\variable{n}$ and center
    $\variable{x}_t$, translates by $\Delta\variable{x}$ under
    environment illumination. \textbf{(b)} The EPIs
    $\variable{L}_t(s,u)$ and $\variable{L}_{t+1}(s,u)$ are correlated
    ($\star$), the loss minimum revealing $\variable{x}_{t+1}$.
    \textbf{(c)} Loss versus recovered translation $x$. Monocular color
    (orange), the classical cue, gives only a shallow dip, the EPIs
    (blue) capture view-dependence but ignore re-illumination, and only
  the relit EPIs (red) drive the loss to zero at the true coordinate.}
  \label{fig:reflection_as_pose_signal}
\end{figure}

%% file: sections/related_work.tex
\section{Related Work}
\label{sec:related}

\textbf{6DoF Pose Estimation and Tracking.} Classical trackers
recover inter-frame motion geometrically, registering point clouds
with \gls{icp} and its colored
variant~\cite{besl1992method,park2017coloredicp}, or solving the
\gls{pnp} problem over matched keypoints with
\gls{ransac}~\cite{lepetit2009ep,fischler1981random}. Learning-based
methods align the observation to a known object model by direct
regression, render-and-compare, and category-level
variants~\cite{xiang_posecnn_2017,wen2020se,labbe2022megapose,
lin2024sam,wang2019normalized}, unified by
FoundationPose~\cite{wen2024foundationpose} into a single estimation
and tracking framework. Model-free trackers such as
BundleSDF~\cite{wen2023bundlesdf}, initialized from LoFTR
matches~\cite{sun2021loftr}, instead build the reference model
online~\cite{wen2023bundlesdf,goncharov2025light}. Whether Lambertian
or view-dependent, these models are treated as invariant to
illumination, capturing the object only under the illumination
present at construction. To our knowledge, no tracker builds a
relightable object model online. We close this gap by recovering the
environment and relighting the model at every frame, keeping the
photometric residual informative as the object moves.

\textbf{Depth Estimation.} Consumer RGB-D cameras match an infrared
pattern between two views~\cite{hirschmuller2008stereo}, assuming a
surface point looks the same from both. Specular surfaces break this
assumption, degrading the recovered depth, an
effect reproduced by our dataset. Light field estimators read
depth from \gls{epi} gradients, defocus and correspondence cues, or
learned cost
volumes~\cite{dansereau2004gradient,tao2013depth,OccCasNet_2024}.
Non-Lambertian extensions add occlusion-aware and specular-removing
aggregation~\cite{wang2015occlusion,tao2015depth}, but all lean on
brightness constancy across views. In our work, we search
for photo-consistent depth in a way that is robust to specular outliers.

\textbf{Novel View Synthesis and Relighting.} Radiance fields and
Gaussian splatting reproduce view-dependent appearance with high
fidelity~\cite{mildenhall2021nerf,kerbl20233d,naylor2026surf},
dynamic extensions carry this to moving
scenes~\cite{pumarola2021d,park2021hypernerf,wu20244d}, and
relightable variants disentangle reflectance from
illumination~\cite{srinivasan2021nerv,gao2024relightable,liu2024bigs,
fan2025spectromotion}. All are fit offline over many views, often per
scene and sometimes under known illumination, which does not suit an
object observed only briefly, in motion, under unknown lighting. We
instead acquire the plenoptic structure they approximate in a single
light field shot, build the object representation feed-forward from
it, and estimate the environment map online from the object's own
reflections.

\textbf{Specular Dynamics.} Shape from specular flow relates the
motion of reflections to surface shape through differential
equations~\cite{canas2009linear,adato2010shape} or sparse reflection
correspondences~\cite{sankaranarayanan2010specular}, but rests on
ideal mirrors, known distant illumination, and smooth or predictable
motion, conditions that an online tracker cannot account for. Our
method handles partially
reflective objects, reconstructs the illumination online, and
accommodates a wide variety of motions.

\textbf{Reflection Separation.} Reflection separation suppresses the
reflected layer as
interference~\cite{sulc2016reflection,shen2023light,li2018robust} or
isolates it as an intrinsic layer in its own
right~\cite{alperovich2016variational}. Operating on raw radiance
without object geometry, these methods separate the layers only in
image space and cannot recover the illumination that produced the
reflection. Our surface light
field ties the reflected
component to the object's depth and normals. The separation yields
the surrounding environment map and supports the relighting that
makes our pose estimation robust.

%% file: sections/method.tex
\section{Method}
\label{sec:method}
\input{figures/method_overview}

\subsection{Reflection as a Pose Signal}
\label{ssec:reflection_signal}
In classical computer vision~\cite{lucas1981iterative}, a moving
point is identified by its color, which holds only for Lambertian
points. A light field image samples the point's view-dependent color,
which helps in a static matching setup~\cite{dansereau2019liff}, yet
once the point moves and is re-illuminated, its angular color
distribution shifts.
\cref{fig:reflection_as_pose_signal} shows a flatland toy example
where the photometric loss of a moving reflective patch zeroes out
only when we relight it under the surrounding environment.

\subsection{Overview}
\label{ssec:overview}
We track a rigid object through a sequence of calibrated light field
frames $\variable{L}_t$, $t = 0, \dots, T-1$, acquired by a static
camera whose subviews have known relative poses. The object is specified
only by a text prompt, and we estimate the relative rigid transform
$\coordframe{t-1}{t}$ between each consecutive pair
$(\variable{L}_{t-1}, \variable{L}_t)$. The object's geometry,
its albedo and the surrounding illumination are all
unknown and recovered along the way. We assume the object is
unoccluded and the illumination static across the sequence, a
distant environment map $\variable{E}$ for the reflections with its
diffuse effect summarized by a single point light. Appearance
follows the dichromatic
reflection model~\cite{shafer1985using}, in which the radiance leaving
a surface point $\variable{x}$ with normal $\variable{n}$ in viewing direction
$\boldsymbol{\omega}_o$ mixes a shaded diffuse color with the environment
sampled along the mirror direction~\cite{blinn1976texture},
\begin{equation}
  \variable{L}_{\mathrm{obs}}(\variable{x}, \boldsymbol{\omega}_o)
  \;=\;
  \alpha \cdot \mathrm{shade}\bigl(\variable{D}(\variable{x}),\,
  \variable{n}\bigr) +
  (1 - \alpha)\,\variable{E}\bigl(\boldsymbol{\omega}_r(\variable{n},
  \boldsymbol{\omega}_o)\bigr),
  \label{eq:brdf}
\end{equation}
where $\mathrm{shade}(\cdot)$ applies Phong diffuse
shading~\cite{phong1998illumination} to the albedo
$\variable{D}(\variable{x})$ under the point light,
$\boldsymbol{\omega}_r(\variable{n},\boldsymbol{\omega}_o) =
2(\variable{n}\!\cdot\!\boldsymbol{\omega}_o)\,\variable{n} -
\boldsymbol{\omega}_o$ mirrors $\boldsymbol{\omega}_o$ about the normal,
and $\alpha \in [0,1]$ is the material's diffuse fraction, with
$\alpha = 1$ purely diffuse and $\alpha = 0$ a perfect mirror.

\cref{fig:method_overview} summarizes our method. Per frame, we
segment the object and lift it to a \emph{surface light field}: a
reflection-robust depth estimate is back-projected to points
$\variable{x}_i$ with normals $\variable{n}_i$ and colors
$\variable{y}_{i, j}$ observed from all subviews
(\cref{ssec:per_frame_representation}). Fitting \cref{eq:brdf} to
these view-dependent samples splits appearance into per-point albedos
$\variable{D}(\variable{x}_i)$ and a single environment map estimate
$\hat{\variable{E}}_t$ shared by all points, with a per-direction
confidence $\variable{C}_t$ recording which parts of the map have been
observed, and calibrates $\alpha$ (\cref{ssec:decomposition}).
Consecutive surface light fields are aligned by a coarse feature match
on the diffuse colors and a point cloud registration, then refined by
transforming the source,
relighting it under $\hat{\variable{E}}_{t-1}$, rendering it, and
descending the photometric residual against the observed frame
(\cref{ssec:pose_alignment}).
Because a moving object keeps mirroring new parts of the scene,
$\hat{\variable{E}}$ and $\variable{C}$ fill in over the sequence,
and the relighting signal sharpens as tracking proceeds.

\subsection{Pose Alignment}
\label{ssec:pose_alignment}
\textbf{Input.} Given the two surface light fields at $t-1$ and $t$
(\cref{ssec:per_frame_representation,ssec:decomposition}), we
estimate the transform $\coordframe{t-1}{t}$ between them. A
frame's surface light field
\begin{equation}
  \mathcal{S}_t = \bigl\{(\variable{x}_i,\, \variable{n}_i,\,
  \variable{D}(\variable{x}_i))\bigr\}_{i=1}^{N},
  \qquad (\hat{\variable{E}}_{t-1},\, \variable{C}_{t-1}),
  \label{eq:slf_repr}
\end{equation}
gathers positions $\variable{x}_i$, normals $\variable{n}_i$, and
albedos $\variable{D}(\variable{x}_i)$. The alignment also
carries the environment map $\hat{\variable{E}}_{t-1}$ accumulated up
to the previous frame and its per-direction observation confidence
$\variable{C}_{t-1}\!\in[0,1]$. To render $\mathcal{S}_t$, we convert
every point into an isotropic, fully opaque 3D
Gaussian~\cite{kerbl20233d} of scale
$\sigma_i \propto \variable{Z}_t(\variable{u}_i)/f$ giving the splat
set $\mathcal{G}_t$ which stores the surface light field. We estimate
$\variable{T} = \coordframe{t-1}{t}$ in two stages
(\cref{fig:method_overview}).

\textbf{Coarse alignment.} To initialize $\variable{T}$, we match the
diffuse central views with LoFTR~\cite{sun2021loftr},
back-project the correspondences through the depth maps, and solve a
relative pose by \gls{ransac}~\cite{fischler1981random}. In parallel we
register the two point clouds with colored
\gls{icp}~\cite{besl1992method,park2017coloredicp}, initialized by their
median shift and a coarse rotation search. We use both, since LoFTR
is stronger on
diffuse surfaces but performs poorly under reflections, whereas \gls{icp}
is reflectivity-invariant given good geometry.

\textbf{Refinement.} Let $\variable{R}$ be the rotation of a candidate
pose $\variable{T}$. To predict the source appearance at
$\variable{T}$, we shade each point's albedo and relight it through the
carried map, giving the per-point color
\begin{equation}
  \variable{c}_i(\variable{T}) =
  \alpha \cdot \mathrm{shade}\bigl(\variable{D}(\variable{x}_i),\,
  \variable{R}\variable{n}_i\bigr)
  + (1-\alpha)\,\hat{\variable{E}}_{t-1}\!\bigl(
    \boldsymbol{\omega}_r(\variable{R}\variable{n}_i,
  \boldsymbol{\omega}_o)\bigr),
  \label{eq:relit}
\end{equation}
with $\mathrm{shade}(\cdot)$ as in \cref{eq:brdf}, evaluated at the
rotated normal. This becomes
the color of
the source Gaussians $\mathcal{G}_{t-1}$, which we splat into the central view,
\begin{equation}
  \hat{\variable{I}}_{\mathrm{pred}}(\variable{T}),\;
  \hat{\variable{Z}}_{\mathrm{pred}}(\variable{T})
  = \mathrm{Render}\bigl(
    \{\,\variable{T}\variable{x}_i,\; \variable{c}_i(\variable{T}),\;
  o_i,\; \sigma_i\,\}\bigr),
  \qquad o_i \equiv 1,
  \label{eq:render}
\end{equation}
with $o_i$ the opacity. To recover the pose, we descend the
photometric residual with a depth anchor,
\begin{equation}
  \variable{T}^\ast = \arg\min_{\variable{T}}\;
  \bigl\| \hat{\variable{I}}_{\mathrm{pred}}(\variable{T})
  - \variable{I}_t \bigr\|_2^2
  + \lambda_z \bigl|
  \hat{\variable{Z}}_{\mathrm{pred}}(\variable{T})
  - \variable{Z}_t \bigr|,
  \label{eq:photo}
\end{equation}
by gradient descent, with $\variable{I}_t$ the observed central view and
$\variable{Z}_t$ its estimated depth. We weight both residuals per
pixel by the rendered
environment confidence $\variable{C}_{t-1}$, deprioritizing
insufficiently observed parts of the environment map.

\subsection{Per-frame Light Field Processing}
\label{ssec:per_frame_representation}
We now recover the quantities used in the alignment process
(\cref{fig:method_overview}).

\textbf{Segmentation.} We use Grounding DINO~\cite{liu2024grounding}
to detect the object's
box in the first frame from a text prompt (\eg
``mustard''), Segment Anything 2~\cite{ravi2025sam} to turn the box
into a mask, and Cutie~\cite{cheng2023putting} to propagate the mask
through the sequence without re-prompting.

\textbf{Depth.} We augment the classic slope formulation of light field
depth~\cite{dansereau2004gradient}, in which a point at depth
$\variable{Z}$, seen at pixel $\variable{u}$ of the central subview,
reappears in subview $\variable{s}_j$ displaced by
$m(\variable{s}_j - \variable{s}_0)$, with $m = f/\variable{Z}$ and $f$ the
focal length. Instead of requiring all subviews to agree on $m$,
which is violated by specular surfaces, we score each slope over only its
best-agreeing subset $\mathcal{V}$~\cite{tao2015depth,wang2015occlusion},
\begin{equation}
  m^\ast(\variable{u}) = \arg\min_{m}\;
  \frac{1}{|\mathcal{V}|}\sum_{j\in\mathcal{V}}
  \Bigl( \ell\bigl(\variable{L}_t(\variable{s}_j,\,\variable{u} +
    m(\variable{s}_j - \variable{s}_0))\bigr)
  - \bar{\ell}_{\mathcal{V}} \Bigr)^{2},
  \label{eq:robust_cost}
\end{equation}
with $\ell$ the luminance and $\bar{\ell}_{\mathcal{V}}$ its mean over
$\mathcal{V}$.

Fully reflective surfaces offer no photo-consistent majority.
However, in practice their silhouette, corners, and reflected edges still match
across subviews, leaving a sparse slope map. To densify
it, we solve the edge-aware weighted least squares
\begin{equation}
  \hat{m} = \arg\min_{m}\;
  \mathcal{L}_{\mathrm{fit}}(m,\,m^\ast)
  + \lambda_m\,\mathcal{L}_{\mathrm{smooth}}(m).
  \label{eq:densify}
\end{equation}
$\mathcal{L}_{\mathrm{fit}}$ anchors the solution to the confident
entries of $m^\ast$, and $\mathcal{L}_{\mathrm{smooth}}$ spreads them
to neighboring pixels without crossing luminance edges of the central
subview. Inverting the dense slopes yields the metric depth of the
current frame, $\variable{Z}_t = f/\hat{m}$.

\textbf{Surface light field initialization.} We unproject every
in-mask pixel into a
point cloud, $\variable{x}_i =
\variable{Z}_t(\variable{u}_i)\,\variable{K}^{-1}\tilde{\variable{u}}_i$,
where $\variable{K}$ is the intrinsics matrix and $\tilde{\variable{u}}_i$ the
homogeneous pixel. To keep normals stable at the silhouette,
where depth differencing fails, we fit them in 3D,
\begin{equation}
  \variable{n}_i = \arg\min_{\|\variable{n}\|=1}
  \sum_{\variable{x}\in\mathcal{N}_k(\variable{x}_i)}
  \bigl(\variable{n}^{\!\top}(\variable{x}-\bar{\variable{x}}_i)\bigr)^{2}
  \quad\text{s.t.}\quad
  \variable{n}^{\!\top}(\variable{e}_0 - \variable{x}_i) > 0,
  \label{eq:normals}
\end{equation}
with $\mathcal{N}_k$ the $k$ nearest neighbors, $\bar{\variable{x}}_i$
their mean, and $\variable{e}_0$ the central camera center
fixing the sign. Reprojecting each point into every subview samples its
directional color profile,
\begin{equation}
  \variable{y}_{i,j} =
  \variable{L}_t\bigl(\variable{s}_j,\,\pi_j(\variable{x}_i)\bigr),
  \qquad
  \boldsymbol{\omega}_{i,j} =
  \frac{\variable{e}_j - \variable{x}_i}
  {\|\variable{e}_j - \variable{x}_i\|},
  \label{eq:slf_colors}
\end{equation}
with $\pi_j$ the projection into subview $j$ and $\variable{e}_j$ its
optical center. These points $\variable{x}_i$, normals
$\variable{n}_i$, and view-dependent colors $\variable{y}_{i, j}$
initialize the frame's surface light field.

\subsection{Diffuse--Specular Decomposition}
\label{ssec:decomposition}
\textbf{Albedo and environment map recovery.}
Given the frame's surface light field $\{\variable{x}_i,
  \variable{n}_i, \variable{y}_{i,j},
\boldsymbol{\omega}_{i,j}\}$, our task is
to restore the points' albedos $\variable{D}(\variable{x}_i)$, the
environment map $\hat{\variable{E}}_t$, and the surface property $\alpha$.
The decomposition is recursive over the sequence. Each frame inherits
$\hat{\variable{E}}_{t-1}$ and the confidence
$\variable{C}_{t-1} \in [0,1]$ from its
predecessor, refines them with the new samples, and passes
$\hat{\variable{E}}_t$ and $\variable{C}_t$ forward. The first frame
starts from an empty map with zero confidence.

Given \textit{a known} $\alpha$, we minimize an augmented version
of~\cref{eq:brdf}
\begin{equation}
  \mathcal{L}_{\mathrm{dec}}
  = \sum_{i,j} w_{i,j}\,
  \Bigl\| \alpha\,\variable{d}_i
  + (1-\alpha)\,\variable{E}\bigl(\boldsymbol{\omega}_r(\variable{n}_i,
  \boldsymbol{\omega}_{i,j})\bigr) - \variable{y}_{i,j} \Bigr\|_2^{2}
  + \mathcal{L}_{\variable{E}}
  + \mathcal{L}_{\variable{d}}
  \label{eq:dec}
\end{equation}
by gradient descent, warm-starting $\variable{E}$ from
$\hat{\variable{E}}_{t-1}$, where $\variable{d}_i$ is the
point's shaded diffuse color. The weight $w_{i,j}$ zeroes samples
that fall outside a subview. $\mathcal{L}_{\variable{E}}$ keeps
$\variable{E}$ in $[0,1]$, anchors
observed bins to $\hat{\variable{E}}_{t-1}$, and applies total
variation weighted by $1 - \variable{C}_{t-1}$, interpolating the
poorly observed environment map pixels from their confident
neighbours. $\mathcal{L}_{\variable{d}}$
smooths $\variable{d}$ with weights that drop across depth and normal
discontinuities.

A final multi-scale pass subtracts the component of
the diffuse image that varies linearly with the back-projected
reflection, stripping leaked texture while preserving local means.
Unshading the result under an assumed point light source,
$\variable{D}(\variable{x}_i) =
\mathrm{shade}^{-1}(\variable{d}_i, \variable{n}_i)$, recovers the
albedos.

The minimizer $\variable{E}$ of \cref{eq:dec} becomes
$\hat{\variable{E}}_t$. Splatting the frame's
reflected rays into the map's bins gives a per-frame coverage
$\tilde{\variable{C}}_t \in [0,1]$, and
$1 - \variable{C}_t = (1 - \variable{C}_{t-1})
(1 - \tilde{\variable{C}}_t)$. $\variable{C}$ only grows as the
motion reveals new parts of the scene.

\textbf{Estimating $\alpha$}. To estimate $\alpha$, we exploit that the diffuse
term of \cref{eq:brdf} is view-independent. Therefore, a point's luminance
spread over the subviews depends only on the specular term,
\begin{equation}
  \nu_i
  = \operatorname{std}_j\, \ell(\variable{y}_{i,j})
  = (1-\alpha)\,\operatorname{std}_j\,
  \ell\bigl(\variable{E}(\boldsymbol{\omega}_r(\variable{n}_i,
  \boldsymbol{\omega}_{i,j}))\bigr).
  \label{eq:alpha_stat}
\end{equation}
We take $\bar{\nu}_t$ as an upper
percentile of the per-point $\nu_i$ accumulated over the sequence.
The spread maps to the diffuse fraction by a linear calibration,
\begin{equation}
  \hat{\alpha}_t
  = \operatorname{clamp}\bigl(1 - \tfrac{\bar{\nu}_t - b}{s},\;
  \alpha_{\min},\, \alpha_{\max}\bigr),
  \label{eq:alpha}
\end{equation}
with offset $b$ and slope $s$ fit once on sequences of known
$\alpha$.

%% file: figures/method_overview.tex
\begin{figure}[!h]
  \centering
  \includegraphics[width=0.86\linewidth]{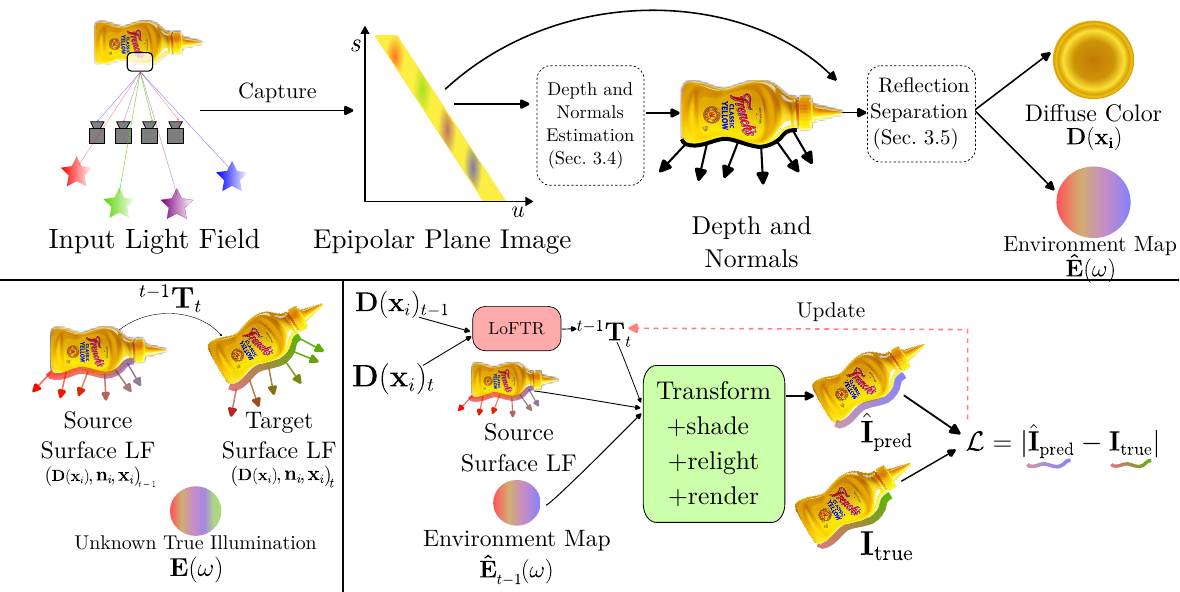}
  \caption[Method overview]{%
    \textbf{Method overview.} \textbf{Top:} each light field frame is
    processed, robustly against reflections, into depth and normals,
    and the view-dependent colors of the surface points are separated
    into per-point albedos $\variable{D}(\variable{x}_i)$ and a shared
    environment map $\hat{\variable{E}}(\boldsymbol{\omega})$
    (\cref{ssec:per_frame_representation,ssec:decomposition}).
    \textbf{Bottom left:} consecutive surface light fields are related
    by the unknown transform $\coordframe{t-1}{t}$; the true
    illumination is never observed, and we substitute it with the map
    $\hat{\variable{E}}_{t-1}$ accumulated over previous frames.
    \textbf{Bottom right:} LoFTR~\cite{sun2021loftr} matches on the
    diffuse colors initialize $\coordframe{t-1}{t}$; the source is then
    transformed, relit under $\hat{\variable{E}}_{t-1}$, and rendered
    into $\hat{\variable{I}}_{\mathrm{pred}}$, whose residual against
  the target updates the pose (\cref{ssec:pose_alignment}).}
  \label{fig:method_overview}
\end{figure}

%% file: sections/dataset.tex
\section{The Multi-Reflectivity Tracking Dataset}
\label{sec:dataset}
To support controlled evaluation of tracking across reflectivity
levels, we introduce a light field object tracking dataset built by
re-rendering YCBInEOAT~\cite{wen2020se}, which preserves
compatibility with prior baselines while making object reflectivity
controllable. Eight
sequences cover four YCB objects (\cref{fig:dataset}b) under various
spatial reorientations. Every sequence comes in two geometry variants, the
original textured mesh and a cube of comparable scale that
isolates reflectance effects from shape effects in ablations. Each
frame is a $5 \times 5$ planar grid of $640 \times 480$ subviews at
$5$~mm baseline whose central subview coincides with the original
YCBInEOAT viewpoint, accompanied by per-subview segmentation masks,
ground truth depth, the 6D object pose, and calibration
(\cref{fig:dataset}a).

\input{figures/figure_dataset_combined}

Each sequence is rendered at four reflectivity levels $r \in \{0.0,
0.5, 0.7, 1.0\}$, sampled more densely toward the specular
end. The diffuse baseline
$\variable{I}_0$ is a Phong~\cite{phong1998illumination} render of
the textured mesh, and the fully reflective extreme $\variable{I}_1$
is a perfect mirror of a fixed distant environment map. Intermediate
levels blend the two per object-pixel, $\variable{I}_r = (1 -
r)\,\variable{I}_0 + r\,\variable{I}_1$, realizing the dichromatic
model of \cref{eq:brdf} with $\alpha = 1 - r$. Geometry, lighting,
and pose are identical across the four levels.

Since consumer RGB-D sensors fail on reflective
surfaces, we additionally pair each frame with a simulated
active-stereo depth map. We render a stereo pair with a projected
infrared dot pattern and estimate depth with
\gls{sgbm}~\cite{hirschmuller2008stereo}, mimicking an Intel
RealSense. As reflectivity increases, the stereo matching cost loses
its minimum and the quality of the recovered depth decreases
(\cref{fig:dataset}c). Rendering, frame filtering, and depth
simulation details are given in the supplementary material.

%% file: figures/figure_dataset_combined.tex
\begin{figure}[!t]
  \centering
  \begin{minipage}[b]{0.47\linewidth}
    \centering
    \includegraphics[width=0.93\linewidth]{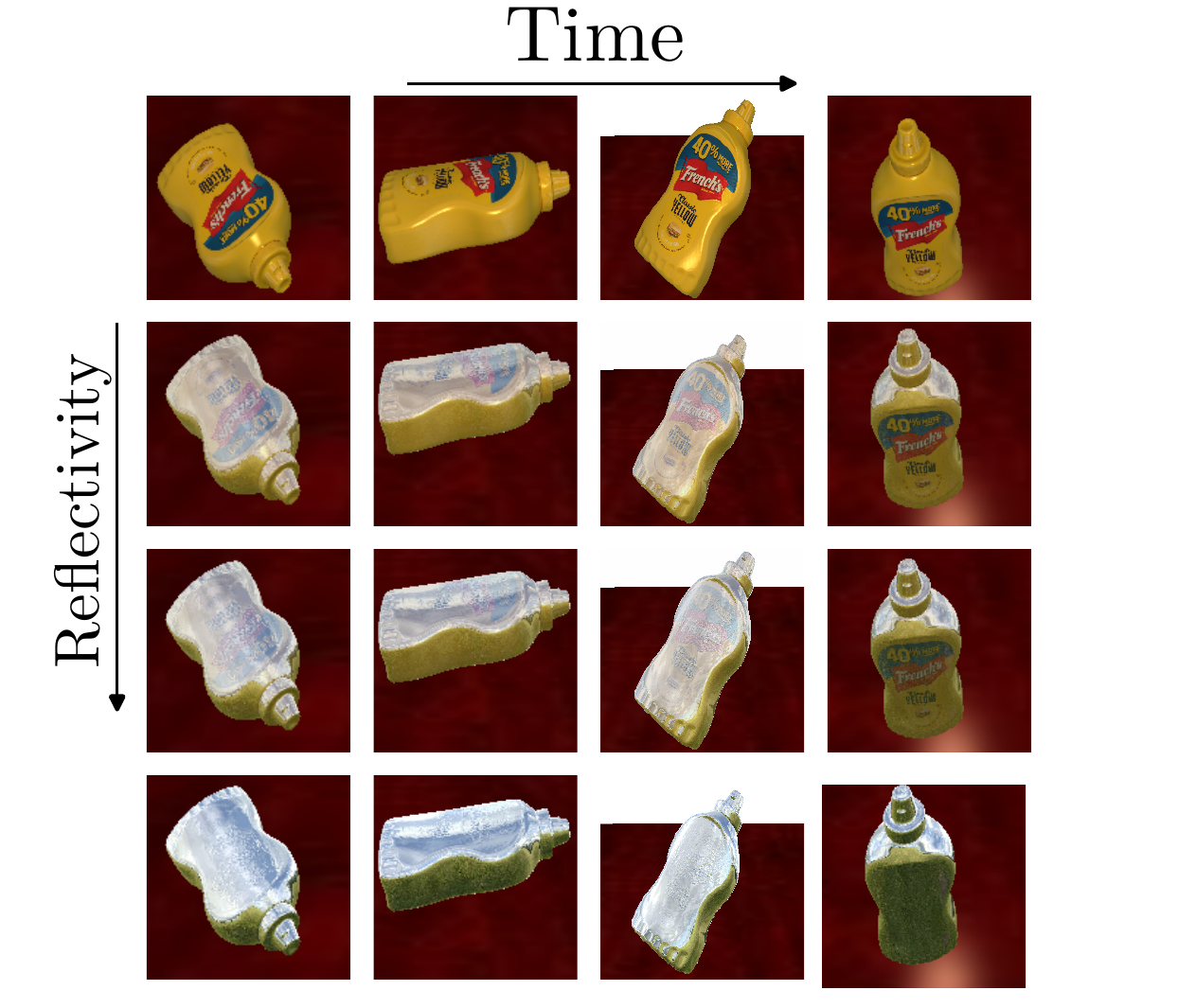}\\[10.3pt]
    {\small (a)}
  \end{minipage}\hfill
  \begin{minipage}[b]{0.47\linewidth}
    \centering
    \includegraphics[width=0.93\linewidth]{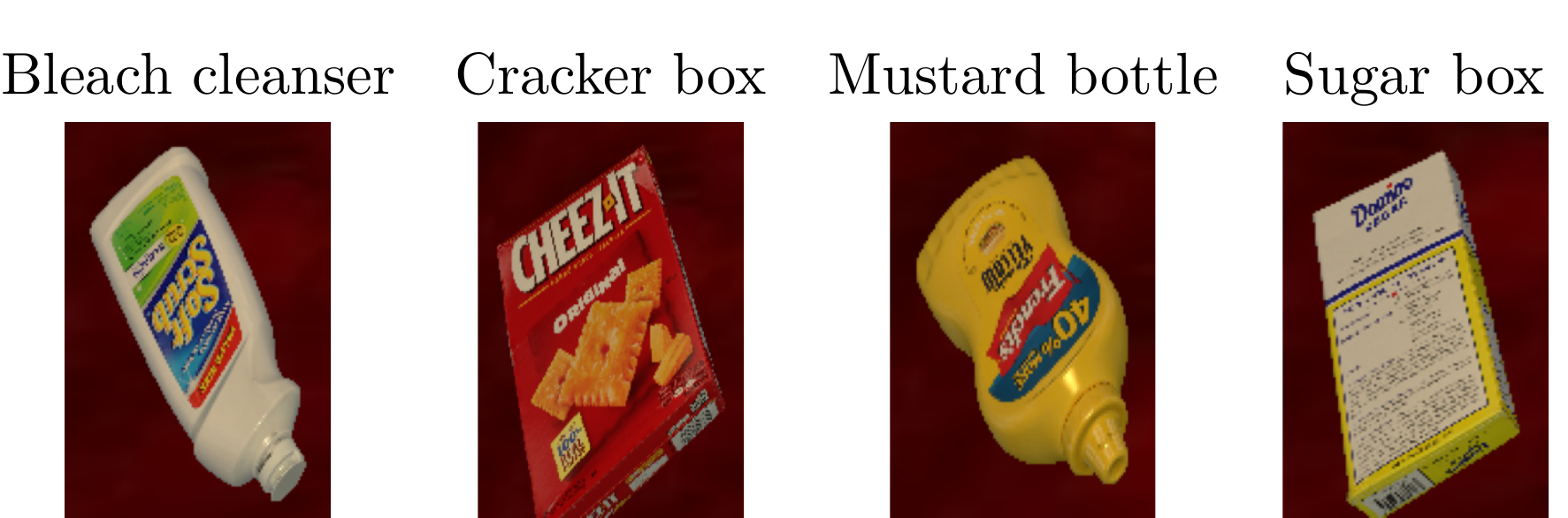}\\
    {\small (b)}\\[-2.5pt]
    \includegraphics[width=0.93\linewidth]{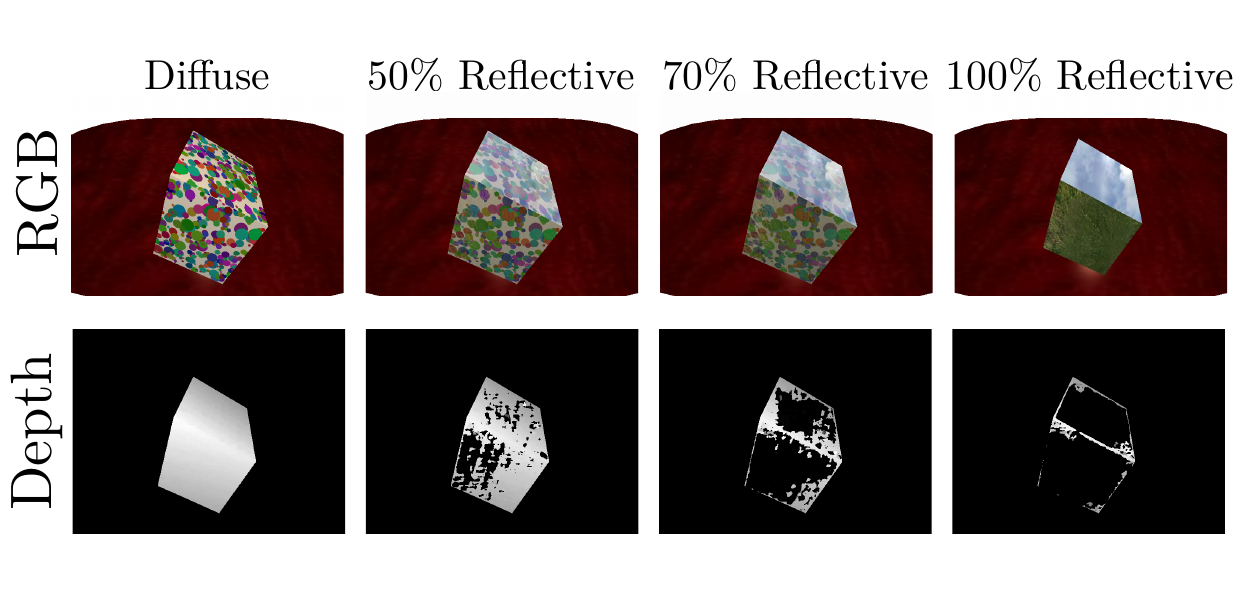}\\[6.5pt]
    {\small (c)}
  \end{minipage}
  \caption[Multi-reflectivity light field tracking dataset]{\textbf{Our
    multi-reflectivity tracking dataset.} (a) The \texttt{mustard0}
    sequence; each frame is available at four reflectivity levels.
    (b) The four YCB objects: bleach cleanser, cracker box, mustard
    bottle, and sugar box. (c) Simulated sensor depth from
    SGBM~\cite{hirschmuller2008stereo} on a rendered stereo pair with a
    projected dot pattern; as reflectivity grows the depth collapses
  into holes, as on an RGB-D camera.}
  \label{fig:dataset}
\end{figure}

%% file: sections/experiments.tex
\section{Experiments}
\label{sec:results}
\subsection{Experimental Setup}
\label{ssec:setup}
We evaluate on the multi-reflectivity dataset of \cref{sec:dataset}.
Every method tracks all eight sequences at all four
reflectivity levels. Following
Xiang \etal~\cite{xiang_posecnn_2017} and Wen
\etal~\cite{wen2024foundationpose}, we report \gls{add} and its
symmetric closest-point variant ADD-S, each summarized as the
\gls{auc} with
threshold $0.1\,d$ for object diameter $d$, as well as the
root-mean-square \gls{ate} and the mean \gls{are}. The baselines
receive the central subview of
each light field, operating as standard monocular RGB-D
methods, while ours receives the full light field. To make the
evaluation invariant to the performance of each baseline's
segmentation, we provide all the methods with ground truth object
masks. Each method runs on the depth it
would have at deployment: the baselines on the simulated sensor depth
and ours on the depth it estimates from the light field. In the
supplementary material, we additionally provide the whole evaluation
on ground truth
depth.

We also record, and release with the dataset, two real-world light
field sequences of one object with an EPIModule camera: a mustard
bottle, plain and then foil-wrapped, moved by a UR5e robot arm through
46 frames each. Ground truth comes from
the robot kinematics through a hand-eye calibration, refined per frame
by aligning the object silhouette.

\subsection{Implementation Details}
\label{ssec:implementation}
The plane sweep (\cref{ssec:per_frame_representation}) scores $192$
inverse-depth hypotheses over $[0.3, 6.0]\,$m and keeps the
best-agreeing $60\%$ of subviews (\cref{eq:robust_cost}). The
diffuse--environment fit (\cref{eq:dec}) runs $300$ Adam steps per
frame, and photometric refinement (\cref{eq:photo}) runs $60$ steps. A drift
guard caps the frame-to-frame pose update on near-diffuse objects. Its
thresholds and the remaining hyperparameters are listed in the
supplementary material.

\subsection{Baselines}
\label{ssec:baselines}
\input{figures/figure_adds_vs_reflectivity_float}
All methods, including ours, are initialized from the ground truth
pose at frame~0. \textbf{PnP} matches \gls{sift}
keypoints between consecutive frames and solves the \gls{pnp}
problem~\cite{lepetit2009ep} with
\gls{ransac}~\cite{fischler1981random}. Colored
\textbf{ICP}~\cite{park2017coloredicp} registers
consecutive masked point clouds. \textbf{LoFTR}~\cite{sun2021loftr}
back-projects detector-free matches through the depth map and fits
the relative pose by \gls{ransac}. These three carry no appearance model.
\textbf{FoundationPose}~\cite{wen2024foundationpose} runs in its
model-free setup, refining photometrically against one neural
reference model built offline from pre-captured views per object and
reflectivity level (shown in the supplementary material).
\textbf{BundleSDF}~\cite{wen2023bundlesdf} learns a neural object
\gls{sdf} online while tracking, kept drift-free by a global
pose-graph optimization.

\subsection{Quantitative Results}
\label{ssec:quantitative_results}

\input{tables/quantitative_main}
\input{tables/real_captures}
\input{figures/figure_trajectory_projected_float}
\input{figures/figure_representation_float}

On the object sequences ours is the only method stable across
reflectivity (\cref{fig:adds_vs_reflectivity},
\cref{tab:main_quantitative}). On diffuse surfaces we trail
FoundationPose and BundleSDF, which use a pre-captured model and global
pose-graph optimization respectively, techniques not used in our
method, and LoFTR only marginally in rotation. At full reflectivity
our method is best on every metric, with the largest gain in
rotation, where photometric refinement is strongest. The baselines
degrade according to the cues they rely on. PnP and LoFTR rest on RGB
feature matching assuming diffuse surfaces. ICP relies on both
geometric and appearance cues, and its tracking quality also degrades.
FoundationPose is limited by its degrading reference
model and the corrupted sensor depth. BundleSDF,
once reflection dominates appearance, is left with too few stable
points to reconstruct from.

On the textured cube (\cref{fig:adds_vs_reflectivity}) our
method does not lead. For perfectly planar objects the distinction
between geometry and appearance remains
ambiguous~\cite{roth2006specular}. Even so, we stay ahead of
BundleSDF at the highest reflectivity, while FoundationPose benefits
from its precomputed object model. The full cube numbers are in the
supplementary material.

Our whole pipeline runs at about 1 fps on an
NVIDIA RTX~6000 Ada (\cref{tab:runtime}), well below real time,
though six times faster
than BundleSDF, the only other tracker that builds its object model
online, while the remaining baselines run at 12 to 34 fps.

On both of the captured sequences (\cref{tab:real_captures}) ours is
best or tied
on every metric. BundleSDF is the only baseline that
stays on the object on both, FoundationPose on the plain bottle only,
and ours cuts BundleSDF's rotation error by roughly a third on each.

\subsection{Qualitative Results}
\label{ssec:qualitative_results}

\cref{fig:trajectory_projected} follows the methods on the fully
reflective sugar box and on the two captured sequences. On the sugar
box (\cref{fig:trajectory_projected}a), LoFTR is deceived by the
reflections and drifts early. FoundationPose drifts once a change of
viewing angle moves the reflections abruptly. BundleSDF stays close
throughout, and our method tracks equally well without its pose-graph
optimization or FoundationPose's pre-captured model, with photometric
refinement correcting the orientation. On the captured sequences
(\cref{fig:trajectory_projected}b,c), LoFTR loses the bottle on both,
FoundationPose on the foil-wrapped one, BundleSDF's silhouette slightly drifts
on the foil, and ours stays aligned.

\cref{fig:representation_qual} shows the representation used
by photometric refinement at $r{=}0.7$. The recovered normals
are reasonable but noisy, and the relit reconstruction is plausible
with low residual. The diffuse and reflected components separate
cleanly on the cube and sugar box but leak into each other on the
bleach and mustard objects, as expected from an underconstrained
decomposition.

\subsection{Ablation Studies}
\label{ssec:ablation_studies}

\input{tables/ablation_runtime}

\cref{tab:ablation} removes one component at a time under our
estimated light field depth. Replacing the light field depth with the
simulated sensor depth is the most damaging change, confirming that
sensor depth cannot be trusted on specular surfaces. Photometric
refinement and reflection separation each cut translation and rotation error by
roughly a quarter, the latter also letting the coarse
frontend match on the recovered diffuse image rather than the raw
reflective one. Predicted masks in place of ground truth ones cost
only a little accuracy. The supplementary material extends this
ablation across reflectivity, both geometries, and the captured sequences.

%% file: figures/figure_adds_vs_reflectivity_float.tex
\begin{figure}[tbp]
  \centering
  \includegraphics[width=0.72\columnwidth]{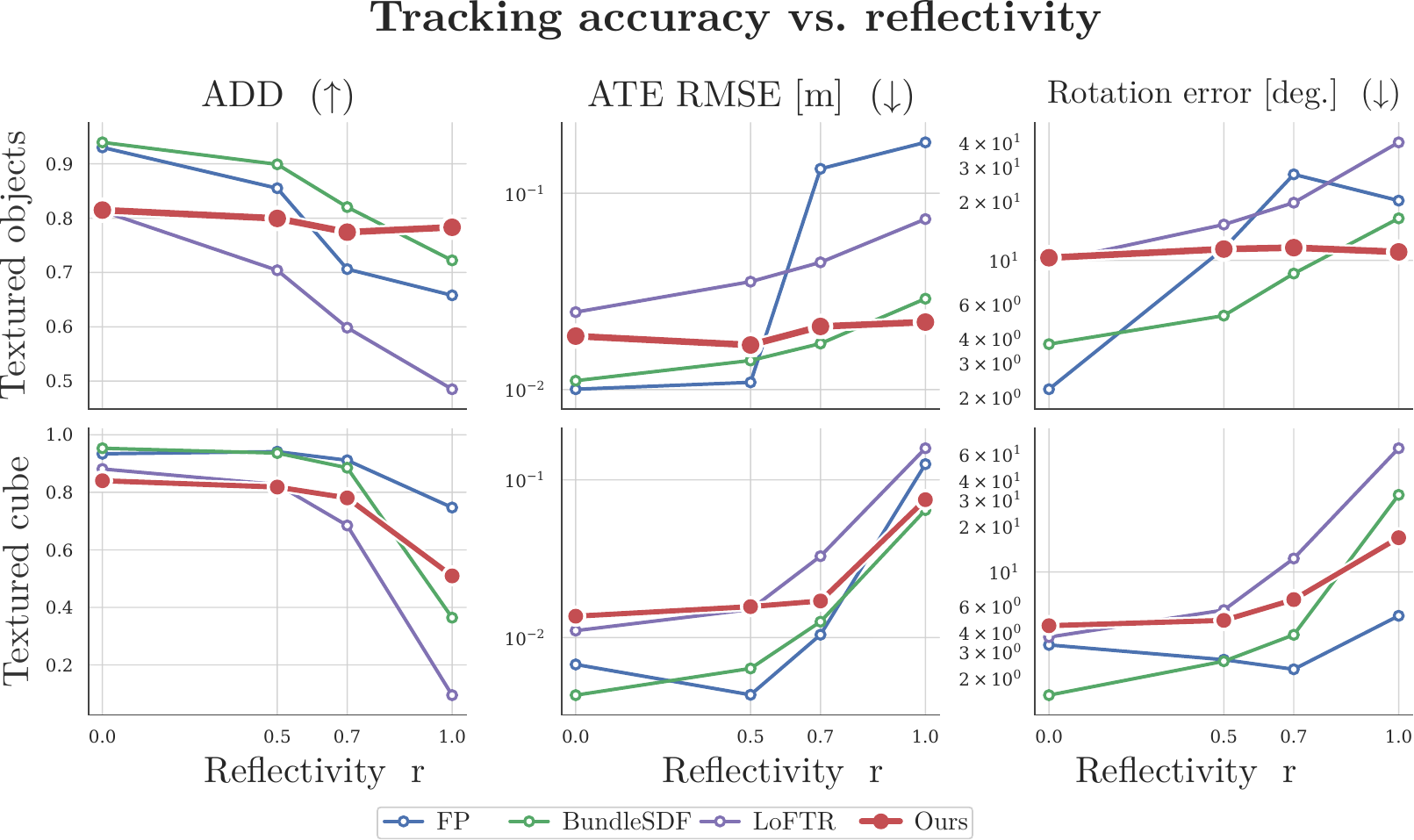}
  \caption[Tracking accuracy versus reflectivity]{\textbf{Tracking accuracy
    versus object reflectivity~$r$}, for the objects (top) and cube
    (bottom), averaged over the eight sequences, reported as ADD (AUC), ATE, and
    ARE; \cref{tab:main_quantitative} gives ADD-S. Ours (red) stays flat
    on the objects while the baselines fall off with reflectivity; on
    the severely underconstrained planar cube the advantage does not
  hold.}
  \label{fig:adds_vs_reflectivity}
\end{figure}

%% file: tables/quantitative_main.tex
\begin{table}[tbp]
  \caption[Average tracking performance on the object
  sequences]{\textbf{Average tracking performance} over the 8 object
    sequences per reflectivity level~$r$ (FP: FoundationPose, BSDF:
    BundleSDF). Ours is the only method stable across the range and
    best on every metric at full reflectivity. Best in \textbf{bold},
    second in \textit{italic}. ADD is given in the supplementary
  material.}
  \label{tab:main_quantitative}
  \centering
  \scriptsize
  \setlength{\tabcolsep}{3pt}
  \renewcommand{\arraystretch}{0.95}
  \begin{tabular}{l cccc cccc cccc}
    \toprule
    & \multicolumn{4}{c}{ADD-S~$\uparrow$}
    & \multicolumn{4}{c}{ATE\,[m]~$\downarrow$}
    & \multicolumn{4}{c}{ARE\,[$^\circ$]~$\downarrow$} \\
    \cmidrule(lr){2-5}\cmidrule(lr){6-9}\cmidrule(lr){10-13}
    $r$ & $0.0$ & $0.5$ & $0.7$ & $1.0$
    & $0.0$ & $0.5$ & $0.7$ & $1.0$
    & $0.0$ & $0.5$ & $0.7$ & $1.0$ \\
    \midrule
    PnP
    & 0.704 & 0.341 & 0.165 & 0.139
    & 0.267 & 0.254 & 1.698 & 1.692
    & 24.8 & 45.7 & 95.4 & 65.4 \\
    ICP
    & 0.794 & 0.566 & 0.442 & 0.370
    & 0.046 & 0.116 & 0.183 & 0.205
    & 27.5 & 44.2 & 56.8 & 68.6 \\
    LoFTR
    & 0.896 & 0.835 & 0.802 & 0.659
    & 0.025 & 0.035 & 0.044 & 0.074
    & 9.8 & 15.2 & 19.7 & 40.0 \\
    FP
    & \textit{0.953} & \textbf{0.937} & 0.733 & 0.691
    & \textbf{0.010} & \textbf{0.011} & 0.133 & 0.182
    & \textbf{2.2} & 11.5 & 27.4 & 20.2 \\
    BSDF
    & \textbf{0.954} & \textit{0.936} & \textbf{0.912} & \textit{0.857}
    & \textit{0.011} & \textit{0.014} & \textbf{0.017} & \textit{0.029}
    & \textit{3.7} & \textbf{5.2} & \textbf{8.6} & \textit{16.4} \\
    \textbf{Ours}
    & 0.898 & 0.893 & \textit{0.881} & \textbf{0.885}
    & 0.019 & 0.017 & \textit{0.021} & \textbf{0.022}
    & 10.3 & \textit{11.4} & \textit{11.6} & \textbf{11.0} \\
    \bottomrule
  \end{tabular}
\end{table}

%% file: tables/real_captures.tex
\begin{table}[!t]
  \caption[Tracking on the captured sequences]{\textbf{Tracking on the
    captured sequences}, two sequences of one mustard bottle, plain and
    foil-wrapped. ADD-S as AUC, ATE in m, ARE in degrees. Best in
    \textbf{bold}, second in \textit{italic}. FoundationPose registers
  on every frame, its tracker being built for smaller steps.}
  \label{tab:real_captures}
  \centering
  \scriptsize
  \setlength{\tabcolsep}{4pt}
  \renewcommand{\arraystretch}{0.95}
  \begin{tabular}{lcccccc}
    \toprule
    & \multicolumn{3}{c}{mustard bottle}
    & \multicolumn{3}{c}{same bottle, foil-wrapped} \\
    \cmidrule(lr){2-4}\cmidrule(lr){5-7}
    Method & ADD-S~$\uparrow$ & ATE\,[m]~$\downarrow$
    & ARE\,[$^\circ$]~$\downarrow$
    & ADD-S~$\uparrow$ & ATE\,[m]~$\downarrow$
    & ARE\,[$^\circ$]~$\downarrow$ \\
    \midrule
    PnP & 0.322 & 0.136 & 31.2 & 0.205 & 0.269 & 128.4 \\
    ICP & 0.135 & 0.193 & 103.2 & 0.207 & 0.172 & 80.7 \\
    LoFTR~\cite{sun2021loftr} & 0.449 & 0.103 & 108.3
    & 0.353 & 0.113 & 99.9 \\
    FoundationPose~\cite{wen2024foundationpose} & 0.868 & 0.038 & 41.1
    & 0.462 & 0.108 & 129.3 \\
    BundleSDF~\cite{wen2023bundlesdf}
    & \textit{0.921} & \textbf{0.013} & \textit{19.9}
    & \textbf{0.941} & \textit{0.011} & \textit{11.2} \\
    \textbf{Ours}
    & \textbf{0.925} & \textbf{0.013} & \textbf{14.0}
    & \textbf{0.941} & \textbf{0.009} & \textbf{6.9} \\
    \bottomrule
  \end{tabular}
\end{table}

%% file: figures/figure_trajectory_projected_float.tex
\begin{figure}[!t]
  \centering
  \begin{subfigure}[t]{0.48\textwidth}
    \centering
    \includegraphics[width=\linewidth,trim=0 0 0
    27pt,clip]{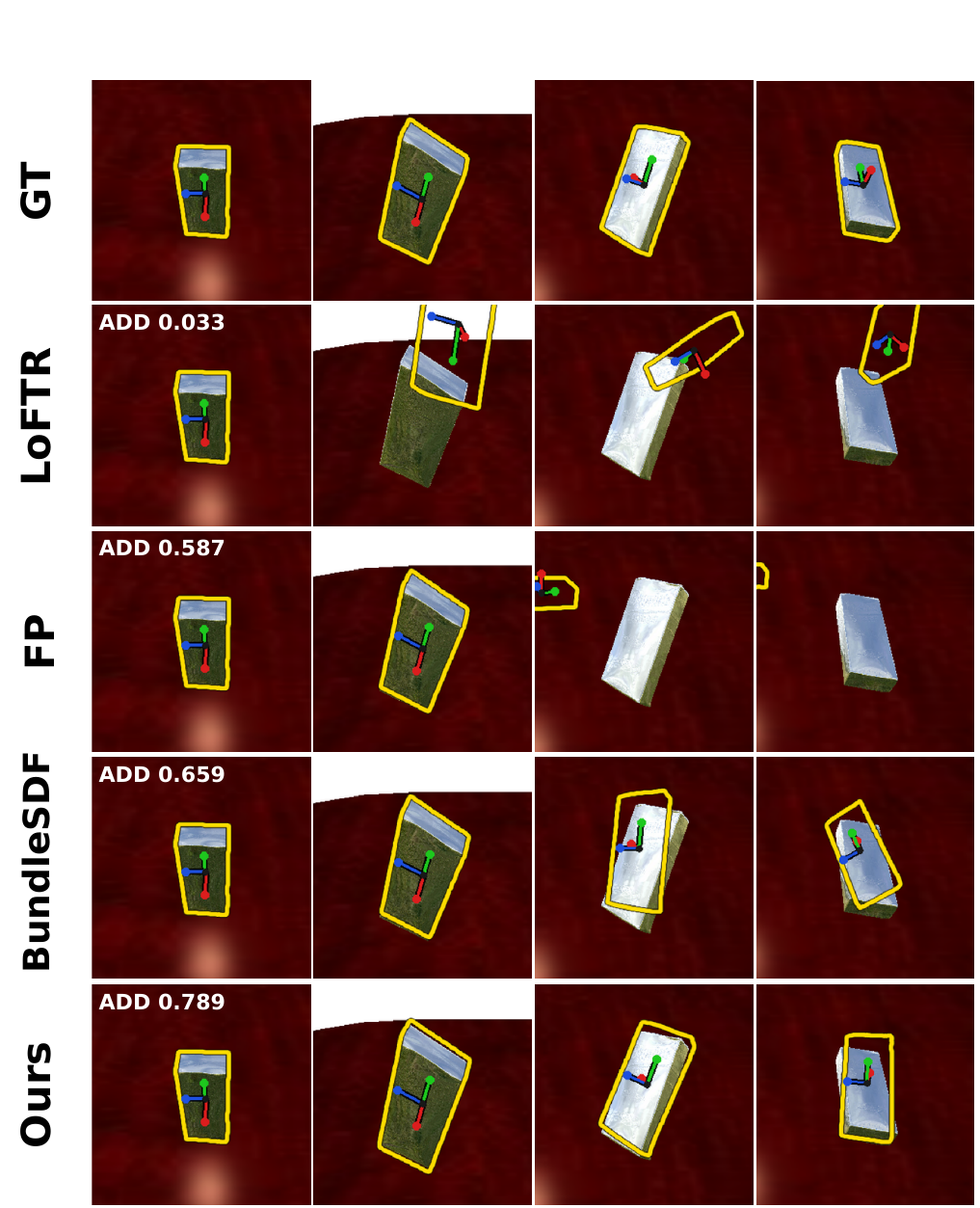}
    \caption{Sugar box, $r{=}1.0$}
  \end{subfigure}\hfill
  \begin{subfigure}[t]{0.48\textwidth}
    \centering
    \includegraphics[width=\linewidth,trim=0 0 0
    27pt,clip]{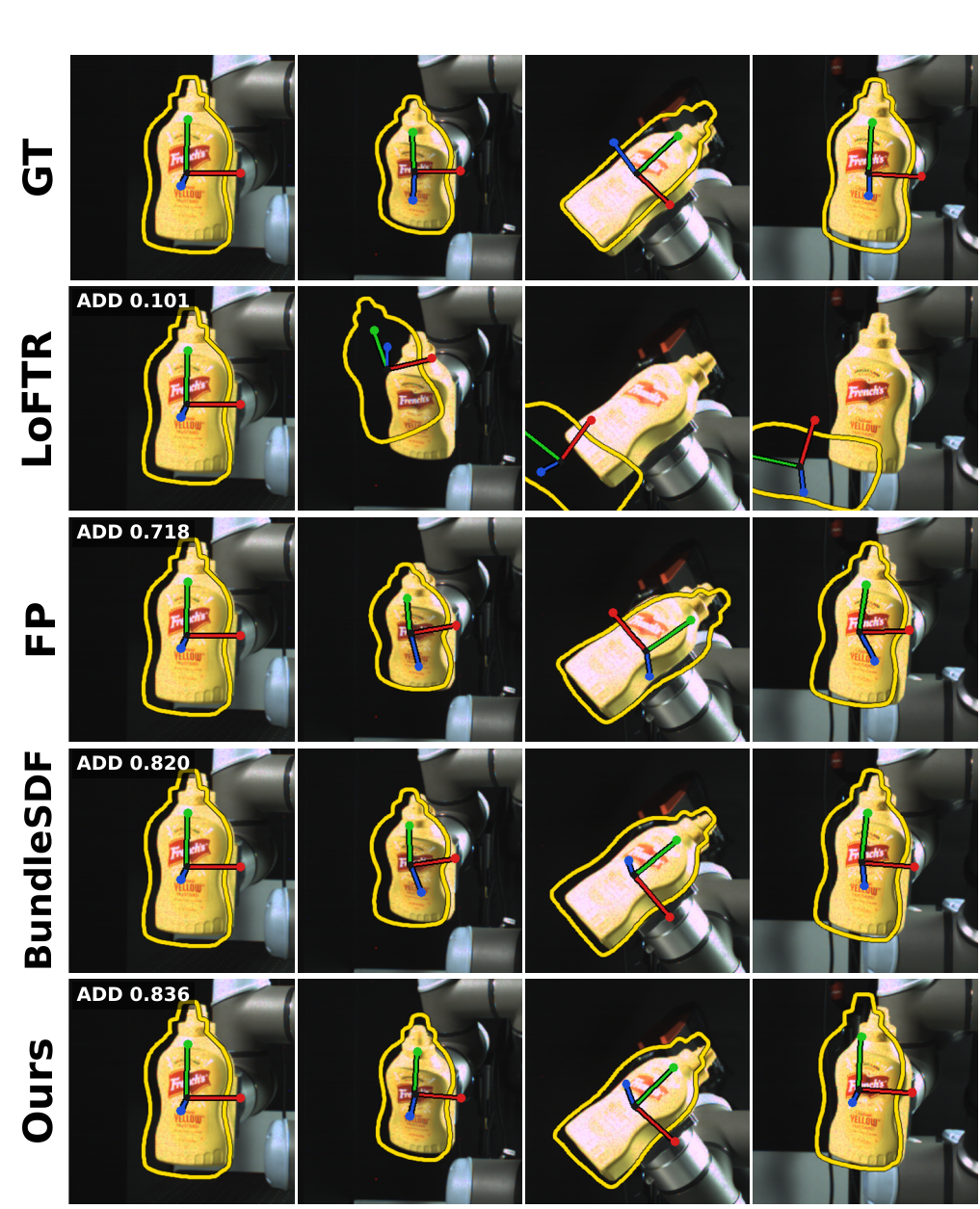}
    \caption{Captured, plain bottle}
  \end{subfigure}\\[6pt]
  \begin{minipage}[t]{0.48\textwidth}
    \vspace{0pt}
    \begin{subfigure}{\linewidth}
      \centering
      \includegraphics[width=\linewidth,trim=0 0 0
      27pt,clip]{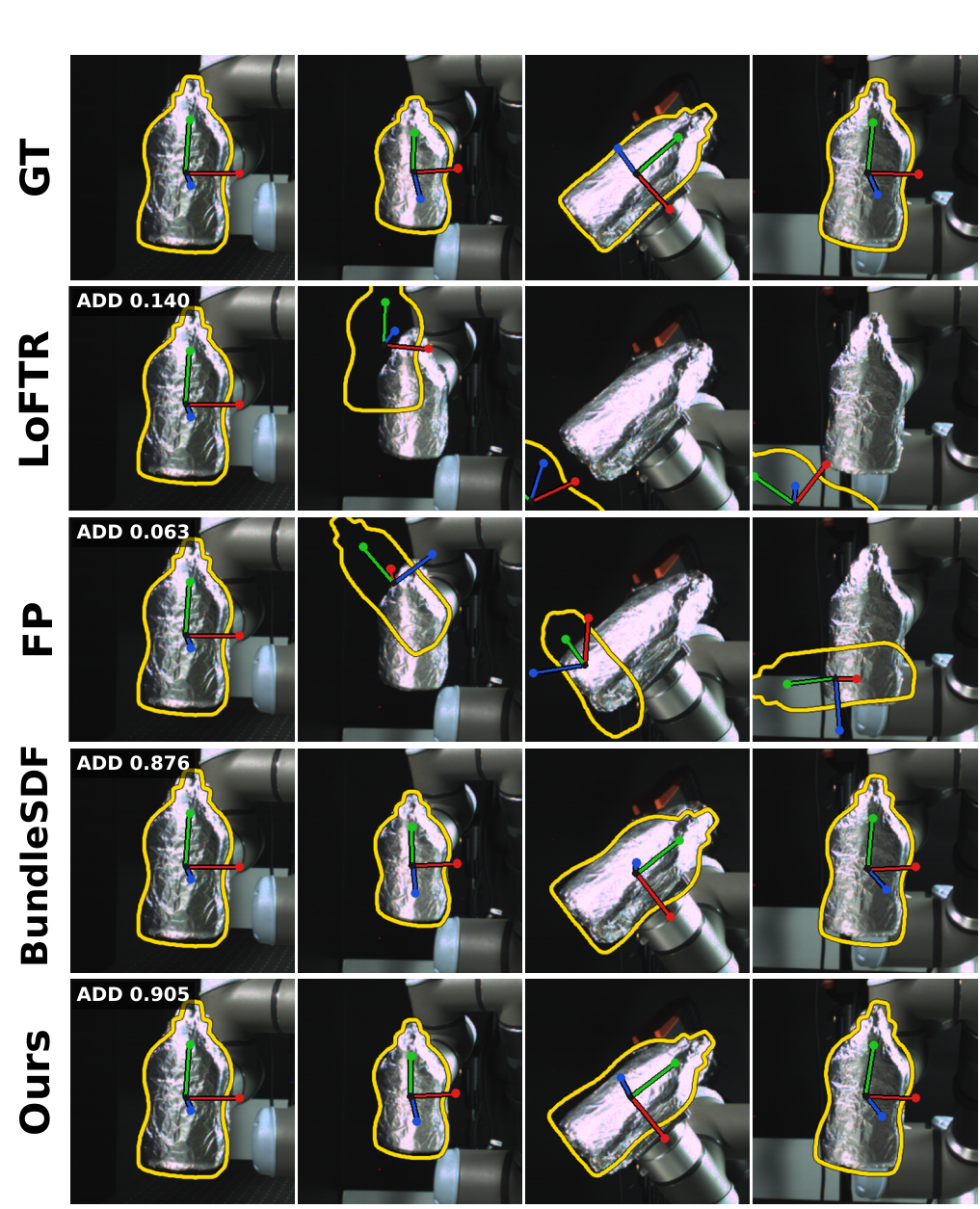}
      \caption{Captured, foil-wrapped}
    \end{subfigure}
  \end{minipage}\hfill
  \begin{minipage}[t]{0.48\textwidth}
    \vspace{0pt}
    \caption[Per-method pose estimates]{\textbf{Per-method pose
      estimates} on (a) the fully specular sugar box
      (\textit{yalehand0}) and the real-world captures of the mustard bottle,
      (b) plain and (c) foil-wrapped. Each row is one method, time
      running left to right; the yellow outline is the estimated
      silhouette and the axes the object frame. Our method corrects the
    orientation throughout.}
    \label{fig:trajectory_projected}
  \end{minipage}
\end{figure}

%% file: figures/figure_representation_float.tex
\begin{figure}[!t]
  \centering
  \includegraphics[width=0.82\textwidth]{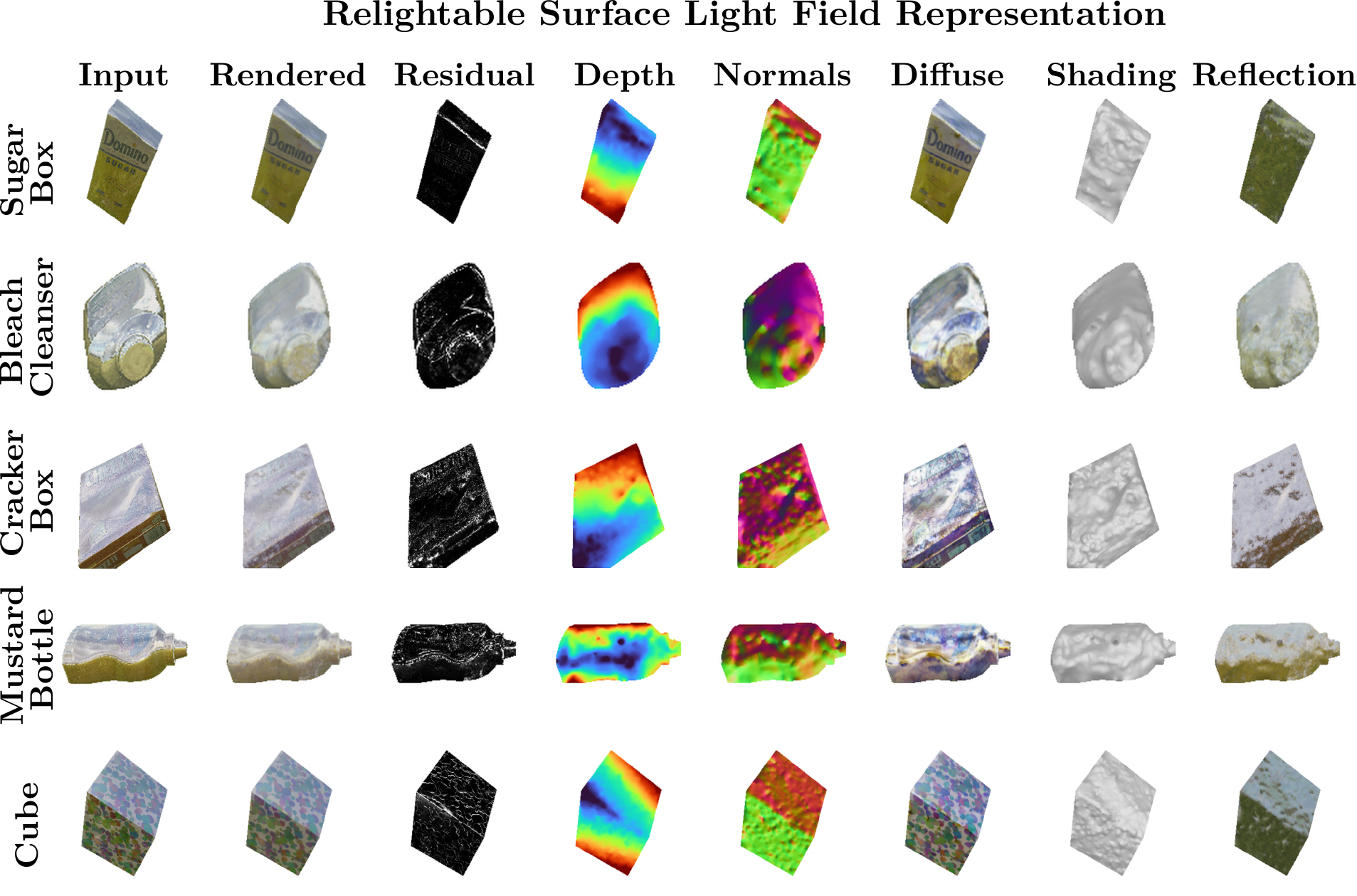}
  \caption[Relightable surface light field
  representation]{\textbf{Relightable surface light field
    representation}, recovered feed-forward from one light field
    capture on estimated depth at $r{=}0.7$. Rows are objects; columns
    show the recovered geometry, the diffuse--environment
  decomposition, and the relit render.}
  \label{fig:representation_qual}
\end{figure}

%% file: tables/ablation_runtime.tex
\begin{table}[!t]
  \begin{minipage}[t]{0.59\textwidth}
    \caption[Component ablation at full reflectivity]{\textbf{Component
      ablation on the objects at full reflectivity} ($r{=}1.0$), one
      component removed per row from \textbf{Ours}; the last row
    substitutes sensor depth.}
    \label{tab:ablation}
  \end{minipage}\hfill
  \begin{minipage}[t]{0.375\textwidth}
    \caption[Throughput and memory across methods]{\textbf{Per-frame
      throughput} and peak GPU memory, RTX~6000 Ada, fully reflective
    sequence. PnP runs on the CPU.}
    \label{tab:runtime}
  \end{minipage}

  \begin{minipage}[t]{0.59\textwidth}
    \centering
    \scriptsize
    \setlength{\tabcolsep}{3pt}
    \renewcommand{\arraystretch}{0.95}
    \begin{tabular}[t]{lcccc}
      \toprule
      Variant & ADD-S & ADD & ATE & ARE \\
      & $\uparrow$ & $\uparrow$ & [m]\,$\downarrow$
      & [$^\circ$]\,$\downarrow$ \\
      \midrule
      Ours w/ predicted masks & 0.878 & 0.769 & 0.023 & 11.6 \\[1.52pt]
      \textbf{Ours} & \textbf{0.885} & \textbf{0.783} & \textbf{0.022}
      & \textbf{11.0} \\[1.52pt]
      $-$ photometric refinement & 0.856 & 0.691 & 0.031 & 14.3 \\[1.52pt]
      $-$ reflection separation & 0.854 & 0.685 & 0.031 & 15.3 \\[1.52pt]
      $-$ light field depth & 0.661 & 0.424 & 0.075 & 30.3 \\[1.52pt]
      \bottomrule
    \end{tabular}
  \end{minipage}\hfill
  \begin{minipage}[t]{0.375\textwidth}
    \centering
    \scriptsize
    \setlength{\tabcolsep}{3pt}
    \renewcommand{\arraystretch}{0.95}
    \begin{tabular}[t]{lcc}
      \toprule
      Method & FPS & Peak GPU \\
      & $\uparrow$ & [GB]\,$\downarrow$\vphantom{[$^\circ$]} \\
      \midrule
      PnP & 34.2 & -- \\
      ICP & 11.7 & 0.9 \\
      LoFTR & 20.1 & 12.9 \\
      FoundationPose & 19.8 & 2.5 \\
      BundleSDF & 0.16 & 14.7 \\
      \textbf{Ours} & 1.03 & 12.9 \\
      \bottomrule
    \end{tabular}
  \end{minipage}
\end{table}

%% file: sections/conclusion.tex
\section{Conclusion}
\label{sec:conclusion}
We presented a \gls{dof} tracker that treats the reflection on a specular
object as a signal for pose. From one light field capture per frame it estimates
depth and normals, separates appearance into diffuse colors and a
shared environment map accumulated online, and tracks by minimizing a
photometric residual against this model relit at each candidate pose,
with no pre-captured mesh or offline reconstruction.

On our multi-reflectivity dataset, ours is the only method
that holds its accuracy as every baseline degrades with increasing
reflectivity. At full reflectivity it leads on every metric, most
clearly in rotation. On diffuse objects we trail the strongest
baselines, and on the geometrically underconstrained planar cube we do
not lead. Two captured sequences add a real-world check, where ours is
best or tied on every
metric. By reasoning about the motion of reflections, this work extends
reliable pose estimation to the reflective objects that current
systems treat as a failure mode.

\textbf{Limitations.} Real-world evaluation covers only two sequences
of one object; broader validation is future work. We assume a static
environment map at
infinity, model appearance with the dichromatic reflection model
rather than spatially varying \glspl{brdf}, do not handle transparent
objects, and carry hyperparameters that may require re-tuning for
other setups. We also solve geometry, intrinsic decomposition, and
motion largely in sequence. Solving them jointly from the light field
and consolidating the per-frame estimates into a single global
relightable model are natural next steps.

%% file: sections/acknowledgements.tex
\section*{Acknowledgments}
This work was supported in part by funding from the Ford Motor
Company and through the NVIDIA Academic Grant Program.

%% file: supplementary.tex
\renewcommand{\topfraction}{0.85}
\renewcommand{\bottomfraction}{0.6}
\renewcommand{\textfraction}{0.07}
\renewcommand{\floatpagefraction}{0.8}
\setcounter{topnumber}{3}
\setcounter{totalnumber}{5}

\suppressfloats[t]
\subsection*{A. Dataset Details}
\label{ssec:supp_dataset}

\begin{sloppypar}
  \textbf{Sequences.} Eight YCBInEOAT~\cite{wen2020se} sequences are
  re-rendered: \path{bleach_hard_00_03_chaitanya}, \path{bleach0},
  \path{cracker_box_reorient}, \path{cracker_box_yalehand0},
  \path{mustard_easy_00_02}, \path{mustard0},
  \path{sugar_box_yalehand0}, and \path{sugar_box1}.
\end{sloppypar}

\textbf{Rendering.} Each frame is rendered with
PyTorch3D~\cite{ravi2020accelerating}, the textured mesh placed at its
ground truth pose in front of a fixed textured background. Ground truth
depth is taken from the z-buffer of the central subview, and
per-subview masks are read from the rasterizer before the background is
composited. \cref{fig:dataset_comparison} sets our re-renderings beside
the original captures.

\input{figures/figure_dataset_comparison_float}

\textbf{Frame filtering.} Since source sequences contain long stationary
stretches, a frame is accepted only if its pose differs from the last
accepted frame by more than $8$~cm in translation or $0.1$~rad in
rotation. \cref{fig:dataset_hists} shows the frame counts before and
after filtering and the distribution of the retained motion.

\input{figures/figure_hists_float}

\textbf{Reflectivity.} The mirror render computes per-pixel reflection
directions from interpolated surface normals and samples them from a
lat-long environment map shared across all sequences. We place the
map at infinity and omit near-field
interreflections. The diffuse--mirror blend is performed in
linear sRGB, with the background left at its diffuse appearance
throughout.

\textbf{Depth simulation.} Each frame is rendered as a horizontal
stereo pair at a $55$~mm baseline, comparable to that of a RealSense. A
sparse binary dot pattern is projected from a virtual emitter rigidly
attached to the left camera, modulated by surface normal incidence and
distance falloff, and added to a low-strength passive infrared base
image. Depth is recovered from the \gls{sgbm} disparity $d$ as
$Z = f b / d$, aligned pixel-for-pixel with the central subview of the
light field. \cref{fig:depth_quality} quantifies how it degrades with
reflectivity.

\input{figures/figure_depth_quality_float}

\FloatBarrier
\subsection*{B. Implementation Details}
\label{ssec:supp_hyperparams}

\cref{tab:hyperparams} lists the remaining hyperparameters of our
method, grouped by pipeline stage.

\input{tables/hyperparams}

\FloatBarrier
\subsection*{C. Baselines}
\label{ssec:supp_baselines}

\cref{tab:baselines_features} summarizes the properties of the
evaluated trackers. Ours is the only one that relights its model.

\input{tables/baselines_features}

\begin{figure}[!htbp]
  \centering
  \includegraphics[width=\columnwidth]{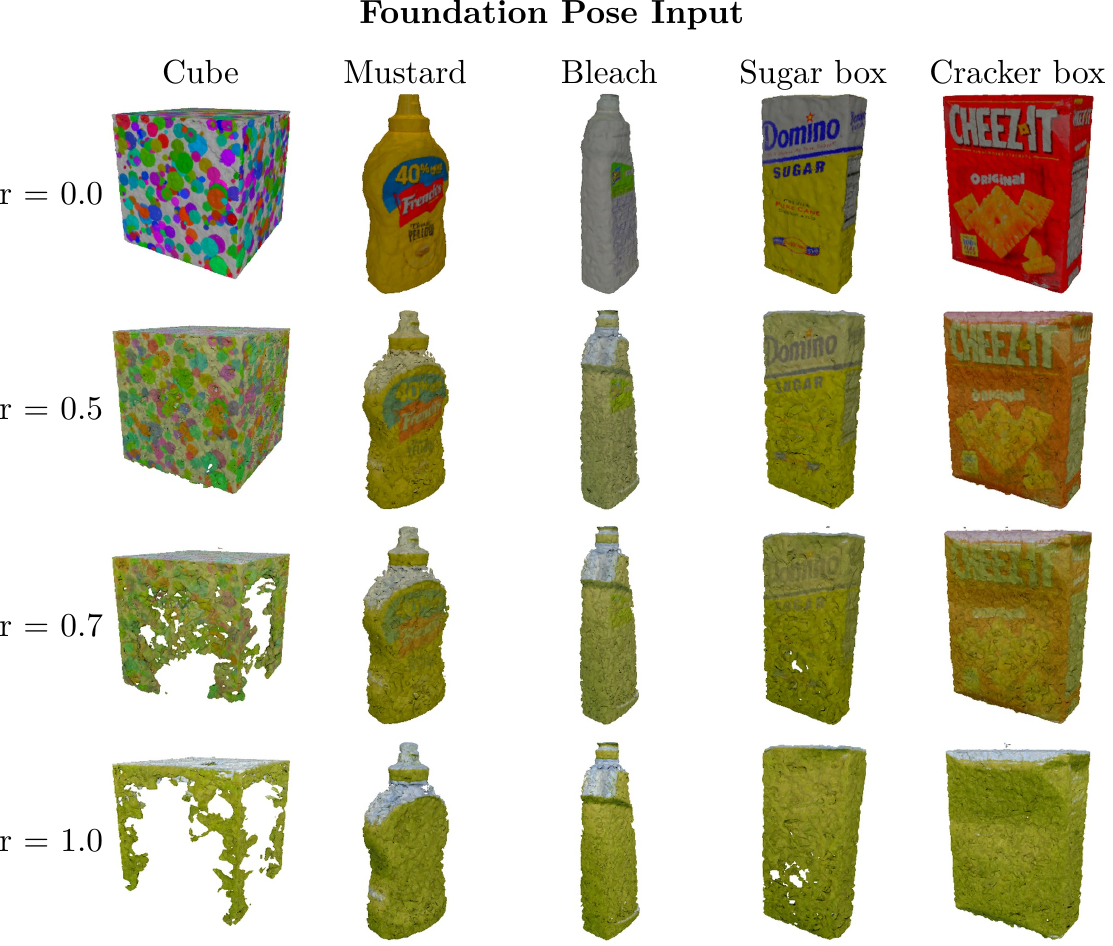}
  \caption[FoundationPose reference models across
  reflectivity]{\textbf{FoundationPose reference models across
    reflectivity.} One neural reference model per object (columns) and
    reflectivity level (rows). As the surface turns specular, holes
    open where the geometry can no longer be recovered, the reflected
    environment is baked into the surface, and the print detail the
  tracker matches against disappears.}
  \label{fig:fp_input}
\end{figure}

\textbf{FoundationPose reference models.} FoundationPose's accuracy
depends on its offline reference model, shown in \cref{fig:fp_input}.
The models are faithful on diffuse surfaces but degrade with
reflectivity, which is one cause of FoundationPose's drop on reflective
objects.

\FloatBarrier
\subsection*{D. Additional Quantitative Results}
\label{ssec:supp_quantitative}

\textbf{Full evaluation grid.} \cref{tab:supp_unified} reports all
metrics, including ADD, over both geometries, both depth regimes, and
all four reflectivity levels, separating the cost of a tracker's depth
source from that of reflective appearance. On the cube our accuracy
holds until full reflectivity, where the planar faces act as mirrors
and our light field depth becomes ambiguous. Given ground truth depth
the cube accuracy largely recovers, placing the cause in the depth
estimate rather than the appearance.

\input{tables/supp_unified}

\textbf{Per-object results at full reflectivity.}
\cref{tab:per_object_hardest} breaks the full-reflectivity results
down per object. Only BundleSDF and our method succeed on every
object, and on the near-planar sugar box, the hardest case, ours leads
all baselines.

\input{tables/per_object_hardest}

\textbf{Depth versus appearance.} Reflections corrupt both the images
the baselines match on and the depth they consume.
\cref{fig:depth_influence} separates the two effects by re-running the
baselines on ground truth depth. On the objects they then stay near
their diffuse-level accuracy at every reflectivity, so reflections in
the depth channel, not only in the RGB images, cause their
degradation. On the cube, BundleSDF collapses at full reflectivity even
under ground truth depth, left without stable appearance to reconstruct
from.

\input{figures/figure_depth_influence_float}

\begin{figure}[tbp]
  \centering
  \includegraphics[width=\columnwidth]{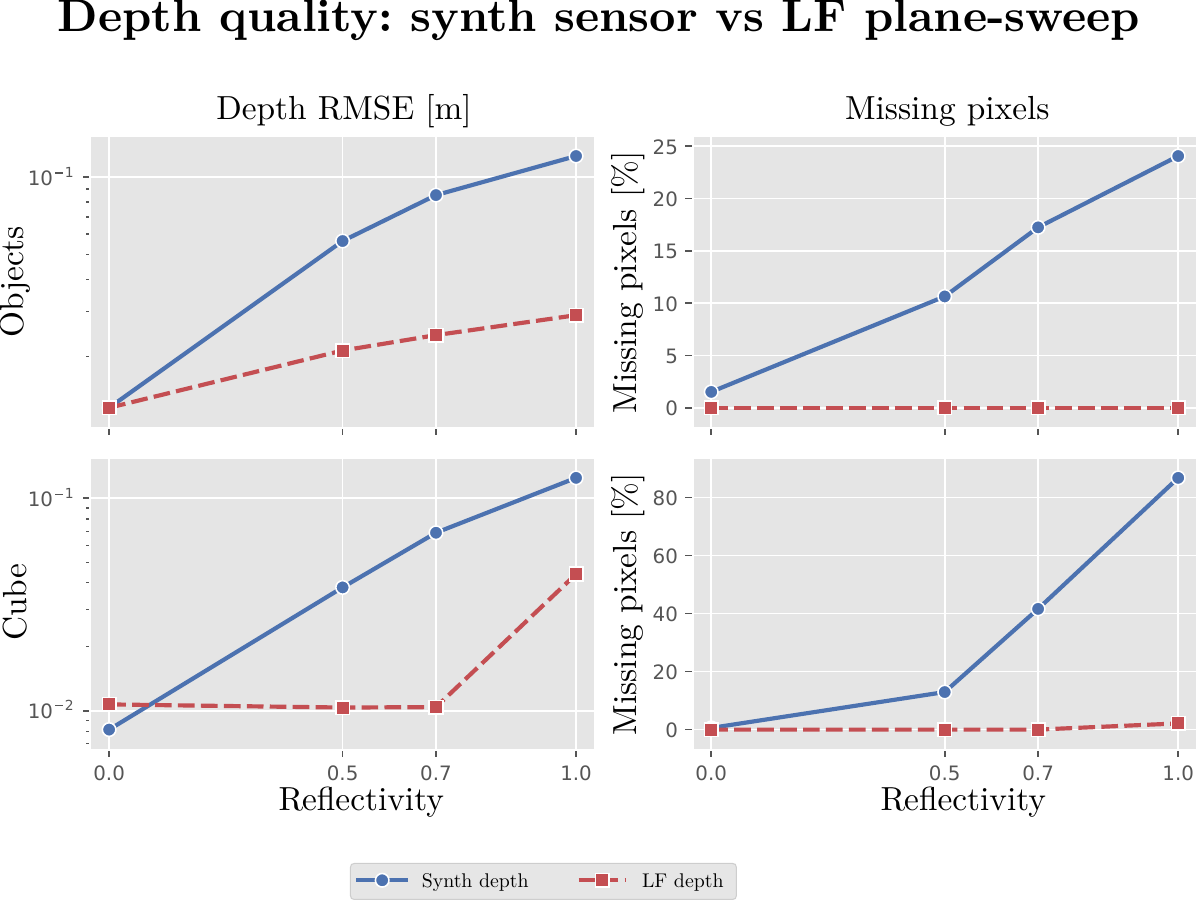}
  \caption[Light field depth versus simulated sensor depth]{\textbf{Depth
    input quality.} Our light field depth (\emph{LF depth}, dashed)
    against the simulated \emph{sensor} depth (solid) versus
    reflectivity $r$, for the objects (top) and cube (bottom): depth
    RMSE (left, log scale) and fraction of missing pixels (right), both
    lower-is-better, averaged over the eight sequences. The sensor depth
    collapses as the surface turns specular while the light field depth
    holds, except on the cube at $r{=}1.0$, where the planar mirror
  makes it ambiguous.}
  \label{fig:depth_sources}
\end{figure}

\textbf{Depth input quality.} \cref{fig:depth_sources} compares the
simulated sensor depth with our light field depth on the same frames.
Ours stays near-metric with almost no missing pixels, except on the
cube at full reflectivity, the depth failure behind our cube results
in \cref{tab:supp_unified}.

\textbf{Reflectivity estimation.} \cref{tab:alpha_final} reports the
online estimate of $\alpha$ at the last frame of every object sequence
as a reflectivity $\hat{r} = 1 - \hat{\alpha}$ against the true $r$.
Diffuse objects are recognized almost exactly. The estimates follow the
true reflectivity in every sequence, with a tendency to underestimate
the most reflective surfaces.

\input{tables/alpha_errors}

\FloatBarrier
\subsection*{E. Additional Qualitative Results}
\label{ssec:supp_qualitative}

\begin{figure}[tbp]
  \centering
  \includegraphics[width=\columnwidth]{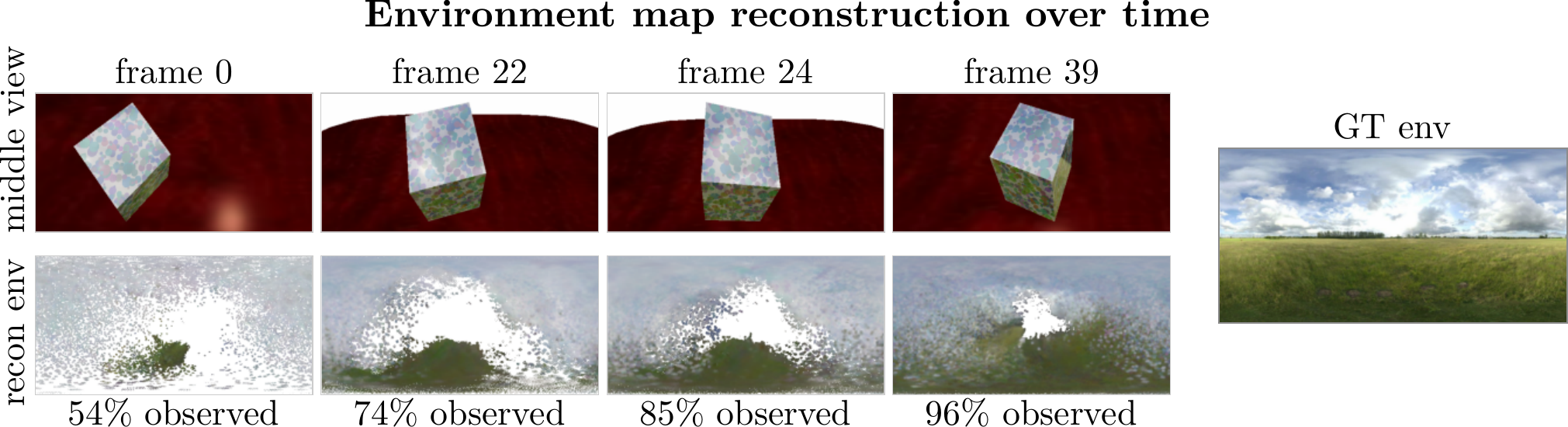}
  \caption[Environment map reconstruction over
  time]{\textbf{Environment map reconstruction over time} on the
    reflective cube ($r{=}0.7$, estimated depth). Top: the central
    subview at four frames. Bottom: the incrementally recovered
    environment map and its observed fraction, with the ground truth
  panorama on the right.}
  \label{fig:envmap_reconstruction}
\end{figure}

\textbf{Environment map reconstruction.} \cref{fig:envmap_reconstruction}
shows the environment map filling in over four frames of the reflective
cube, approaching the ground truth panorama.

\textbf{Reflection separation.} The decomposition is driven by the
estimated surface normals, so it inherits any error in the estimated
geometry. \cref{fig:reflection_separation} compares the separation
under ground truth and estimated depth.

\input{figures/figure_reflection_separation_float}

\FloatBarrier
\subsection*{F. Additional Ablations}
\label{ssec:supp_ablation}

\textbf{Ablation across reflectivity.} \cref{tab:ablation_combined}
extends the component ablation to all four reflectivity levels and
both geometries. Replacing light field depth with sensor depth is the
most damaging change throughout, and photometric refinement and
reflection separation contribute more the more reflective the surface:
negligible on diffuse objects but decisive on the fully reflective
cube, where they separate holding the pose from losing it. Predicted
masks cost little accuracy except on that same hardest case.

\textbf{Ablation on the captured sequences.}
\cref{tab:real_captures_ablation} repeats the component ablation on the
two captured sequences. The ordering matches the synthetic results:
replacing light field depth is by far the most damaging change,
reflection separation carries the rotation gain on the foil, and
photometric refinement there improves position only.

\textbf{Qualitative ablation.} \cref{fig:qualitative_ablation} shows
the ablation as tracking behavior on the fully reflective sugar box,
the hardest object case.

\input{tables/ablation_combined}
\input{tables/real_captures_ablation}
\FloatBarrier
\input{figures/figure_ablation_float}

%% file: figures/figure_dataset_comparison_float.tex
\begin{figure}[tbp]
  \centering
  \includegraphics[width=\columnwidth]{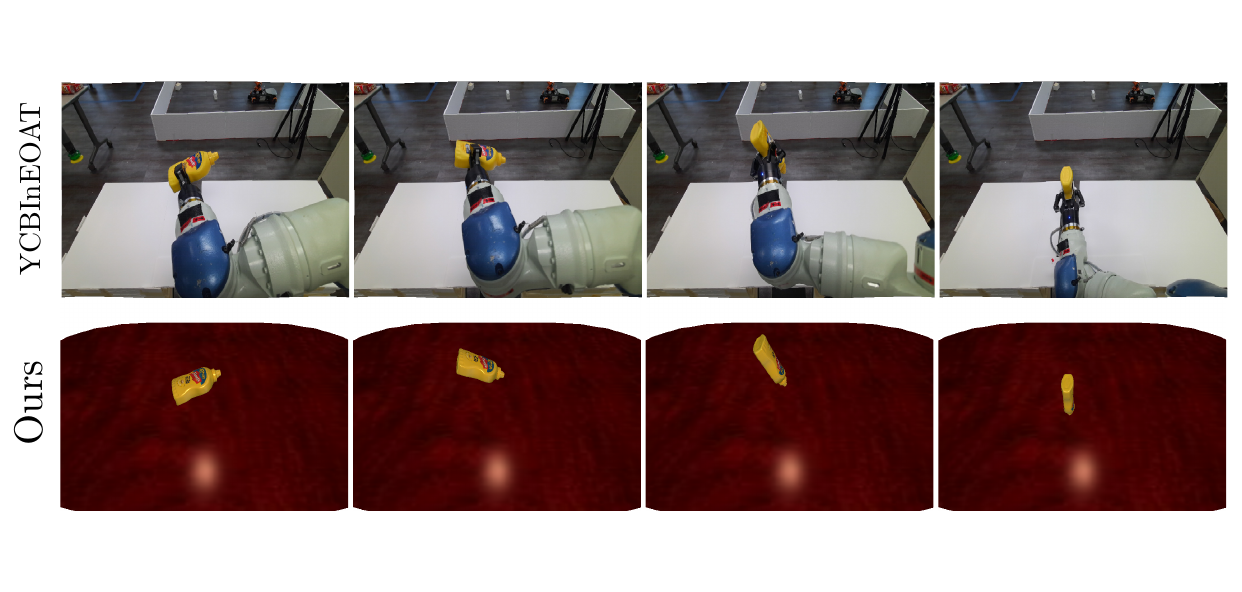}
  \caption[Comparing the simulated light field object
  tracking dataset with its captured counterpart]{\textbf{Dataset
    comparison.} Original YCBInEOAT~\cite{wen2020se} captures (top) and
    our re-rendered central subviews at the same frames (bottom). The
    object follows the same ground truth trajectory in both, while the
    lab scene is replaced by a controlled background so reflectivity
  can be varied without changing the motion.}
  \label{fig:dataset_comparison}
\end{figure}

%% file: figures/figure_hists_float.tex
\begin{figure}[tbp]
  \centering
  \includegraphics[width=\columnwidth]{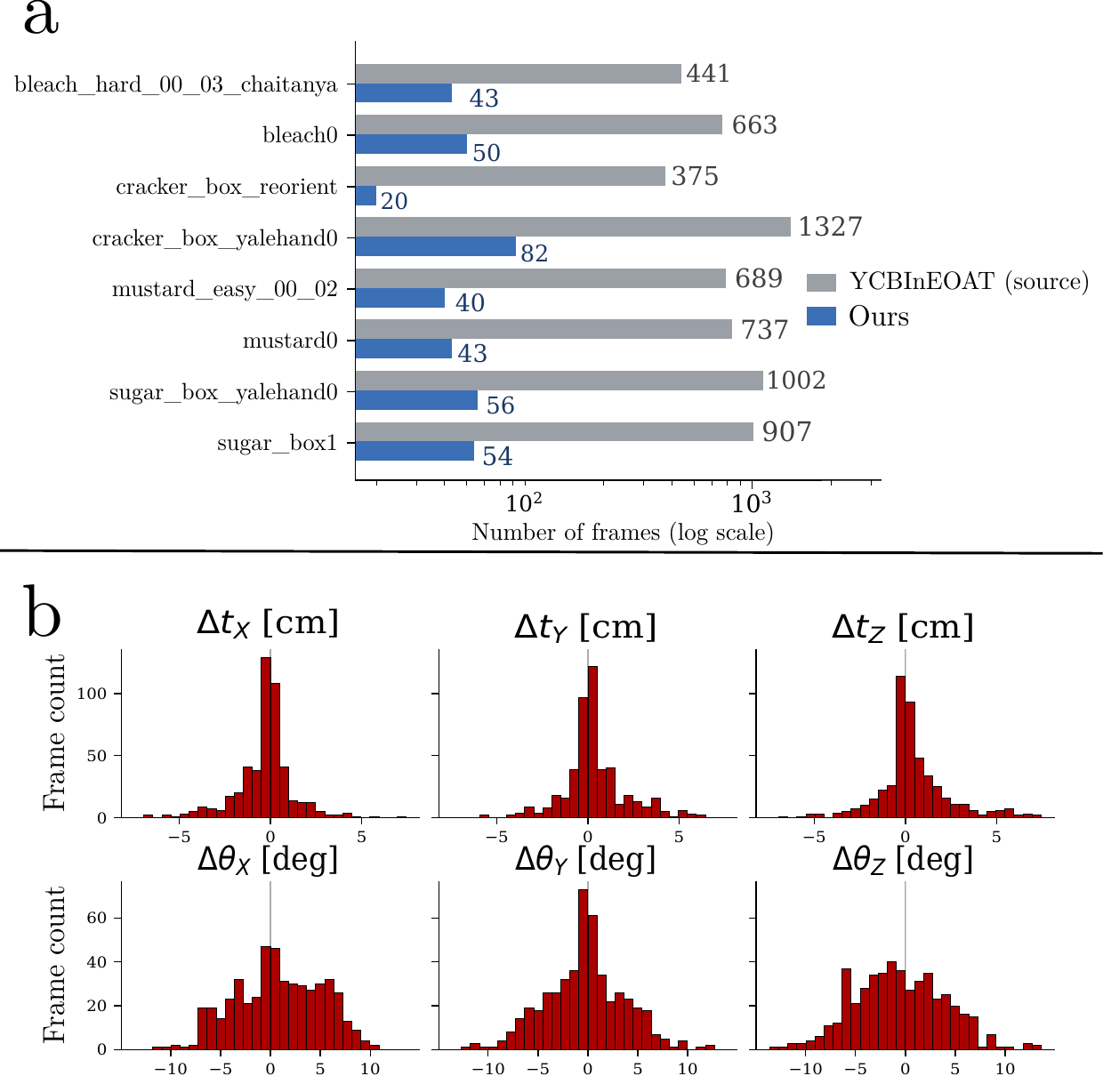}
  \caption[Simulated light field object
  tracking dataset statistics]{\textbf{Dataset composition} compared
    to YCBInEOAT~\cite{wen2020se}. (a) Per-sequence frame counts
    before (gray) and after (blue) motion filtering: filtering removes
    redundant near-stationary frames, reducing the total by roughly an
    order of magnitude while preserving every sequence. (b)
    Distribution of per-frame motion in the filtered dataset across
    the six degrees of freedom: the retained motion stays broad and
  well spread on every axis.}
  \label{fig:dataset_hists}
\end{figure}

%% file: figures/figure_depth_quality_float.tex
\begin{figure}[tbp]
  \centering
  \includegraphics[width=\columnwidth]{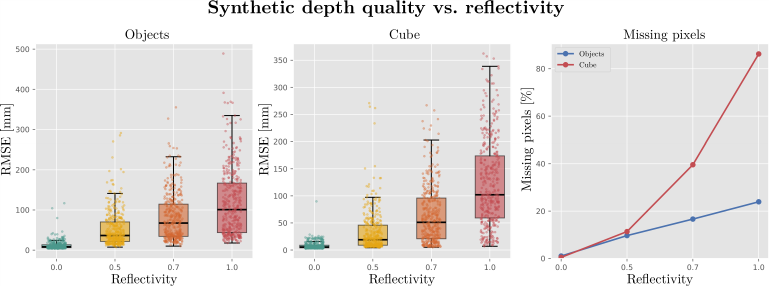}
  \caption[Simulated depth quality as a function of
  reflectivity]{\textbf{Simulated sensor depth versus reflectivity.}
    Per-frame depth RMSE distributions on the objects (left) and cube
    (center), and the mean fraction of missing pixels (right), as
    functions of reflectivity $r$. From $r{=}0$ to $r{=}1$ the median
    RMSE grows roughly tenfold and the cube loses up to $87\%$ of its
  pixels.}
  \label{fig:depth_quality}
\end{figure}

%% file: tables/hyperparams.tex
\begin{table}[!htbp]
  \caption[Remaining method hyperparameters]{\textbf{Remaining
    hyperparameters} of our method, grouped by pipeline stage: depth
    estimation, decomposition, coarse alignment, and drift guard
    (left); photometric refinement, relighting, and shading
  (right).}
  \label{tab:hyperparams}
  \centering
  \footnotesize
  \renewcommand{\arraystretch}{1.0}
  \setlength{\tabcolsep}{4pt}
  \adjustbox{max width=\columnwidth,center}{%
    \begin{tabular}[t]{@{}l l@{}}
      \toprule
      Parameter & Value \\
      \midrule
      \multicolumn{2}{@{}l}{\emph{LF depth}} \\
      matching cost window                & $7$\,px \\
      smoothness weight / edge scale      & $4.0$ / $0.06$ \\
      gradient vs.\ intensity cost weight & $8.0$ \\
      texture-confidence gradient scale   & $0.05$ \\
      confidence exponent                 & $1.0$ \\
      edge-weight floor                   & $0.02$ \\
      conjugate-gradient iterations       & $160$ \\
      mirror reject (quantile / margin)   & $0.10$ / $0.30\,$m \\
      \midrule
      \multicolumn{2}{@{}l}{\emph{Decomposition}} \\
      separation steps per frame          & $300$ \\
      $\alpha$ calibration offset $b$ / slope $s$
      & $2{\times}10^{-3}$ / $24{\times}10^{-3}$ \\
      $\bar{\nu}_t$ percentile (within / across frames) & $30$ / $80$ \\
      $\alpha$ clamp $[\alpha_{\min}, \alpha_{\max}]$ & $[0.03, 0.97]$ \\
      $\alpha$ stability tolerance         & $0.01$ \\
      \midrule
      \multicolumn{2}{@{}l}{\emph{Coarse alignment}} \\
      min.\ LoFTR inliers                 & $12$ \\
      Gaussian scale ($\times$ pixel footprint) & $0.5$ \\
      \midrule
      \multicolumn{2}{@{}l}{\emph{Drift guard}} \\
      rotation cap                        & $10^\circ$ \\
      translation cap                     & $0.015\,$m \\
      enabled above diffuse fraction      & $0.8$ \\
      \bottomrule
    \end{tabular}
    \hspace{10pt}
    \begin{tabular}[t]{@{}l l@{}}
      \toprule
      Parameter & Value \\
      \midrule
      \multicolumn{2}{@{}l}{\emph{Refinement}} \\
      rotation / translation LR           & $5{\times}10^{-3}$ /
      $10^{-4}$ \\
      depth-anchor weight $\lambda_z$     & $3.0$ \\
      unobserved-bin confidence floor     & $0.15$ \\
      render scales                       & $1$ \\
      cosine LR decay                     & off \\
      keep step only if loss improves     & yes \\
      early-stop patience / min.\ rel.\ improvement & $10$ /
      $5{\times}10^{-3}$ \\
      per-frame rollback cap (rot / trans) & disabled \\
      \midrule
      \multicolumn{2}{@{}l}{\emph{Relight stage}} \\
      rotation / translation LR           & $3{\times}10^{-3}$ /
      $2{\times}10^{-4}$ \\
      \midrule
      \multicolumn{2}{@{}l}{\emph{Shading (Phong)}} \\
      light position (camera frame)       & $(0,-0.1,0)$ \\
      ambient / diffuse / specular        & $0.5$ / $(0.5,0.4,0.25)$
      / $(0.5,0.45,0.35)$ \\
      shininess                           & $64$ \\
      \bottomrule
  \end{tabular}}
\end{table}

%% file: tables/baselines_features.tex
\providecolor{passgreen}{HTML}{27AE60}
\providecolor{failred}{HTML}{C0392B}
\providecommand{\gcmark}{\textcolor{passgreen}{\cmark}}
\providecommand{\rxmark}{\textcolor{failred}{\xmark}}
\begin{table}[!htb]
  \centering
  \caption[Properties of our baseline object tracking
  methods]{\textbf{Properties of our baseline object tracking methods.}
    Feature presence (\gcmark) or absence
  (\rxmark) for each evaluated tracker.}
  \label{tab:baselines_features}
  \footnotesize
  \setlength{\tabcolsep}{4pt}
  \begin{adjustbox}{max width=\linewidth}
    \begin{tabular}{lccccc}
      \toprule
      Method & Appearance & Photometric & Pose-graph   & Pre-captured
      & Relighting \\
      & modeling  & refinement  & optimization & views        &            \\
      \midrule
      PnP~\cite{fischler1981random}        & \rxmark & \rxmark &
      \rxmark & \rxmark & \rxmark \\
      ICP~\cite{park2017coloredicp}        & \rxmark & \rxmark &
      \rxmark & \rxmark & \rxmark \\
      LoFTR~\cite{sun2021loftr}                  & \rxmark & \rxmark &
      \rxmark & \rxmark & \rxmark \\
      FoundationPose~\cite{wen2024foundationpose} & \gcmark & \gcmark &
      \rxmark & \gcmark & \rxmark \\
      BundleSDF~\cite{wen2023bundlesdf}          & \gcmark & \gcmark &
      \gcmark & \rxmark & \rxmark \\
      Ours                                       & \gcmark & \gcmark &
      \rxmark & \rxmark & \gcmark \\
      \bottomrule
    \end{tabular}
  \end{adjustbox}
\end{table}

%% file: tables/supp_unified.tex
\begin{sidewaystable}
  \caption[Full evaluation grid across geometry, depth and
  reflectivity]{\textbf{Full evaluation grid} over both geometries
    (\emph{Objects}, \emph{Cube}), both depth regimes, and four
    reflectivity levels~$r$, each value averaging the eight motion
    sequences. \emph{Sensor} is the deployment regime, in which the
    baselines consume the simulated sensor depth and ours the depth it
    estimates from the light field; \emph{GT} gives every method ground
    truth depth. Best in \textbf{bold}, second best in \textit{italic},
    per metric within each block. Ours is the steadiest across
    reflectivity on the objects, while on the planar cube it degrades at
    $r{=}1.0$, where the mirror ambiguity corrupts our estimated
  depth.}
  \label{tab:supp_unified}
  \centering
  \footnotesize
  \setlength{\tabcolsep}{3pt}
  \renewcommand{\arraystretch}{1.1}
  \newcommand{\mh}{ADD-S\,$\uparrow$ &
    ADD\,$\uparrow$ & ATE\,[m]\,$\downarrow$ &
  ARE\,[$^\circ$]\,$\downarrow$}
  \adjustbox{max width=\textheight,center}{%
    \begin{tabular}{@{}c l @{\hskip 7pt} cccc @{\hskip 6pt} cccc
      @{\hskip 12pt} cccc @{\hskip 6pt} cccc @{}}
      \toprule
      & & \multicolumn{8}{c}{Objects}
      & \multicolumn{8}{c}{Cube} \\
      \cmidrule(lr){3-10}\cmidrule(lr){11-18}
      & & \multicolumn{4}{c}{Sensor} & \multicolumn{4}{c}{GT}
      & \multicolumn{4}{c}{Sensor} & \multicolumn{4}{c}{GT} \\
      \cmidrule(lr){3-6}\cmidrule(lr){7-10}\cmidrule(lr){11-14}\cmidrule(lr){15-18}
      $r$ & Method & \mh & \mh & \mh & \mh \\
      \midrule
      \multirow{6}{*}{$r=0.0$}
      & PnP & 0.704 & 0.607 & 0.267 & 24.8 & 0.702 & 0.603 &
      0.193 & 30.1 & 0.899 & 0.881 & 0.024 & 6.2 & 0.818 & 0.799 &
      0.037 & 19.5 \\
      & ICP & 0.794 & 0.590 & 0.046 & 27.5 & 0.882 & 0.774 &
      0.026 & 10.4 & 0.192 & 0.192 & 0.148 & 65.5 & 0.706 & 0.702 &
      0.038 & 19.0 \\
      & LoFTR~\cite{sun2021loftr} & 0.896 & 0.813 & 0.025 & 9.8 &
      0.927 & 0.867 & 0.014 & 7.7 & 0.882 & 0.881 & 0.011 & 3.7 &
      0.923 & \textit{0.923} & \textit{0.006} & 2.4 \\
      & FoundationPose~\cite{wen2024foundationpose} & \textit{0.953}
      & \textit{0.930} & \textbf{0.010} & \textbf{2.2} &
      \textit{0.971} & \textit{0.957} & \textit{0.004} & \textit{2.0}
      & \textbf{0.956} & \textit{0.934} & \textit{0.007} &
      \textit{3.3} & \textit{0.925} & 0.901 & 0.013 & 4.5 \\
      & BundleSDF~\cite{wen2023bundlesdf} & \textbf{0.954} &
      \textbf{0.940} & \textit{0.011} & \textit{3.7} & \textbf{0.980}
      & \textbf{0.977} & \textbf{0.003} & \textbf{1.2} &
      \textit{0.954} & \textbf{0.953} & \textbf{0.004} & \textbf{1.5}
      & \textbf{0.980} & \textbf{0.980} & \textbf{0.002} & \textbf{0.7} \\
      & \textbf{Ours} & 0.898 & 0.815 & 0.019 & 10.3 & 0.924 & 0.855
      & 0.014 & 6.2 & 0.841 & 0.841 & 0.014 & 4.4 & 0.901 & 0.901 &
      0.011 & \textit{2.2} \\
      \midrule
      \multirow{6}{*}{$r=0.5$}
      & PnP & 0.341 & 0.220 & 0.254 & 45.7 & 0.337 & 0.183 &
      0.321 & 61.3 & 0.827 & 0.827 & 0.026 & 5.1 & 0.755 & 0.755 &
      0.050 & 19.6 \\
      & ICP & 0.566 & 0.306 & 0.116 & 44.2 & 0.813 & 0.620 &
      0.044 & 13.0 & 0.367 & 0.363 & 0.067 & 25.3 & 0.634 & 0.597 &
      0.032 & 15.8 \\
      & LoFTR & 0.835 & 0.704 & 0.035 & 15.2 & 0.883 & 0.776 & 0.025
      & 12.0 & 0.828 & 0.827 & 0.015 & 5.6 & 0.888 & 0.888 & 0.008 & 3.6 \\
      & FoundationPose & \textbf{0.937} & \textit{0.855} &
      \textbf{0.011} & 11.5 & \textit{0.966} & \textit{0.947} &
      \textit{0.005} & \textit{2.8} & \textbf{0.952} & \textbf{0.942}
      & \textbf{0.004} & \textit{2.6} & \textit{0.963} &
      \textit{0.936} & \textit{0.004} & 3.7 \\
      & BundleSDF & \textit{0.936} & \textbf{0.899} & \textit{0.014}
      & \textbf{5.2} & \textbf{0.973} & \textbf{0.963} &
      \textbf{0.004} & \textbf{2.0} & \textit{0.938} & \textit{0.937}
      & \textit{0.006} & \textbf{2.6} & \textbf{0.969} &
      \textbf{0.969} & \textbf{0.003} & \textbf{0.8} \\
      & \textbf{Ours} & 0.893 & 0.799 & 0.017 & \textit{11.4} & 0.935
      & 0.879 & 0.012 & 4.3 & 0.819 & 0.819 & 0.016 & 4.8 & 0.880 &
      0.880 & 0.011 & \textit{3.2} \\
      \midrule
      \multirow{6}{*}{$r=0.7$}
      & PnP & 0.165 & 0.099 & 1.698 & 95.4 & 0.264 & 0.143 &
      0.557 & 95.3 & 0.487 & 0.486 & 0.099 & 20.0 & 0.628 & 0.628 &
      0.067 & 19.1 \\
      & ICP & 0.442 & 0.207 & 0.183 & 56.8 & 0.762 & 0.514 &
      0.059 & 14.9 & 0.226 & 0.208 & 0.106 & 52.4 & 0.461 & 0.447 &
      0.052 & 23.2 \\
      & LoFTR & 0.802 & 0.598 & 0.044 & 19.7 & 0.857 & 0.722 & 0.033
      & 14.2 & 0.685 & 0.685 & 0.033 & 12.3 & 0.801 & 0.801 & 0.013 & 6.6 \\
      & FoundationPose & 0.733 & 0.706 & 0.133 & 27.4 &
      \textit{0.962} & \textit{0.939} & \textbf{0.006} & \textit{3.6}
      & \textbf{0.912} & \textbf{0.912} & \textbf{0.010} &
      \textbf{2.3} & \textit{0.961} & \textit{0.935} & \textit{0.004} & 3.6 \\
      & BundleSDF & \textbf{0.912} & \textbf{0.820} & \textbf{0.017}
      & \textbf{8.6} & \textbf{0.965} & \textbf{0.952} &
      \textit{0.007} & \textbf{2.7} & \textit{0.886} & \textit{0.886}
      & \textit{0.013} & \textit{3.9} & \textbf{0.963} &
      \textbf{0.963} & \textbf{0.004} & \textbf{0.8} \\
      & \textbf{Ours} & \textit{0.881} & \textit{0.774} &
      \textit{0.021} & \textit{11.6} & 0.938 & 0.885 & 0.012 & 4.0 &
      0.782 & 0.781 & 0.017 & 6.6 & 0.886 & 0.886 & 0.011 & \textit{2.9} \\
      \midrule
      \multirow{6}{*}{$r=1.0$}
      & PnP & 0.139 & 0.078 & 1.692 & 65.4 & 0.149 & 0.081 &
      0.444 & 79.8 & 0.051 & 0.051 & 2.876 & 90.3 & 0.046 & 0.042 &
      1.246 & 111.1 \\
      & ICP & 0.370 & 0.153 & 0.205 & 68.6 & 0.668 & 0.415 &
      0.084 & 16.8 & 0.086 & 0.085 & 0.304 & 91.2 & 0.195 & 0.190 &
      0.131 & 37.5 \\
      & LoFTR & 0.659 & 0.485 & 0.074 & 40.0 & 0.815 & 0.602 & 0.044
      & 19.4 & 0.097 & 0.095 & 0.159 & 65.2 & 0.079 & 0.074 & 0.307 & 80.7 \\
      & FoundationPose & 0.691 & 0.658 & 0.182 & 20.2 &
      \textit{0.955} & \textit{0.924} & \textit{0.007} & \textit{4.2}
      & \textbf{0.748} & \textbf{0.747} & 0.126 & \textbf{5.1} &
      \textbf{0.918} & \textbf{0.918} & \textbf{0.015} & \textbf{2.1} \\
      & BundleSDF & \textit{0.857} & \textit{0.722} & \textit{0.029}
      & \textit{16.4} & \textbf{0.965} & \textbf{0.949} &
      \textbf{0.005} & \textbf{2.9} & 0.369 & 0.364 & \textbf{0.064}
      & 32.2 & 0.067 & 0.065 & 0.402 & 87.4 \\
      & \textbf{Ours} & \textbf{0.885} & \textbf{0.783} &
      \textbf{0.022} & \textbf{11.0} & 0.931 & 0.864 & 0.013 & 5.0 &
      \textit{0.510} & \textit{0.510} & \textit{0.075} &
      \textit{16.8} & \textit{0.754} & \textit{0.750} &
      \textit{0.020} & \textit{7.1} \\
      \bottomrule
  \end{tabular}}
\end{sidewaystable}

%% file: tables/per_object_hardest.tex
\providecommand{\fail}[1]{#1\,\textsuperscript{\scriptsize\xmark}}
\begin{table}[t]
  \centering
  \caption[Per-object robustness at full reflectivity]{\textbf{Per-object
    ADD}~$\uparrow$ at full reflectivity ($r{=}1.0$), averaged over
    each object's sequences. An object is \emph{lost} (\xmark) when
    mean ADD falls below $0.5$; the last row counts the objects lost.
  Best per row in \textbf{bold}.}
  \label{tab:per_object_hardest}
  \footnotesize
  \setlength{\tabcolsep}{4pt}
  \begin{tabular}{l c c c c c c}
    \toprule
    Object & PnP & ICP
    & LoFTR & FP & BSDF & Ours \\
    \midrule
    Bleach      & \fail{0.080} & \fail{0.147} & 0.554 & 0.525 &
    \textbf{0.810} & 0.795 \\
    Cracker box & \fail{0.056} & \fail{0.086} & 0.525 & \textbf{0.881}
    & 0.526 & 0.736 \\
    Mustard     & \fail{0.129} & \fail{0.289} & 0.771 & \textbf{0.914}
    & 0.909 & 0.861 \\
    Sugar box   & \fail{0.048} & \fail{0.090} & \fail{0.090} &
    \fail{0.311} & 0.643 & \textbf{0.741} \\
    \midrule
    Objects lost~$\downarrow$ & 4 & 4 & 1 & 1 & \textbf{0} &
    \textbf{0} \\
    \bottomrule
  \end{tabular}
\end{table}

%% file: figures/figure_depth_influence_float.tex
\begin{figure}[t]
  \centering
  \includegraphics[width=\columnwidth]{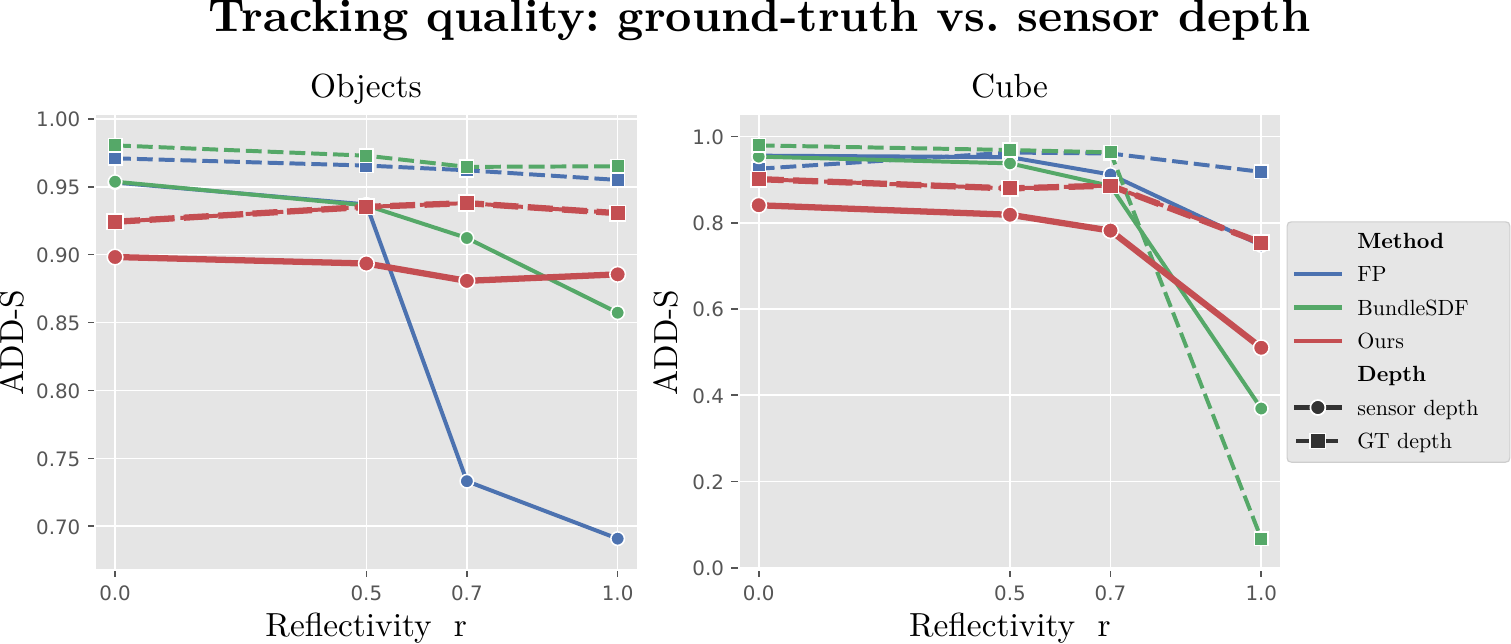}
  \caption[Tracking under ground truth versus deployment
  depth]{\textbf{Tracking under ground truth versus deployment depth.}
    Mean ADD-S over the eight sequences on the objects (left) and cube
    (right), under \emph{ground truth} depth (dashed) and each method's
    deployment depth (solid; simulated sensor depth for the baselines,
    our estimated light field depth for ours). With clean depth the
    baselines barely degrade on the objects, so their collapse under
    sensor depth is caused by depth corruption rather than reflective
    appearance. The cube is the exception: BundleSDF collapses at full
  reflectivity under either depth.}
  \label{fig:depth_influence}
\end{figure}

%% file: tables/alpha_errors.tex
\begin{table}[tbp]
  \caption[Final reflectivity estimates]{\textbf{Final reflectivity
    estimates on the objects}, per sequence and true reflectivity $r$.
    Each entry is the estimated reflectivity $\hat{r} = 1 - \hat{\alpha}$
    at the last frame of the sequence (light field depth). Estimates are
    clamped to $[0.03, 0.97]$.
  }
  \label{tab:alpha_final}
  \centering
  \footnotesize
  \setlength{\tabcolsep}{6pt}
  \begin{adjustbox}{max width=\linewidth}
    \begin{tabular}{lcccc}
      \toprule
      Sequence & $r{=}0.0$ & $r{=}0.5$ & $r{=}0.7$ & $r{=}1.0$ \\
      \midrule
      bleach0 & 0.03 & 0.44 & 0.64 & 0.97 \\
      bleach\_hard\_00\_03\_chaitanya & 0.03 & 0.35 & 0.48 & 0.60 \\
      cracker\_box\_reorient & 0.09 & 0.80 & 0.90 & 0.97 \\
      cracker\_box\_yalehand0 & 0.03 & 0.91 & 0.97 & 0.97 \\
      mustard0 & 0.03 & 0.45 & 0.63 & 0.90 \\
      mustard\_easy\_00\_02 & 0.03 & 0.33 & 0.47 & 0.71 \\
      sugar\_box1 & 0.05 & 0.53 & 0.65 & 0.78 \\
      sugar\_box\_yalehand0 & 0.03 & 0.45 & 0.59 & 0.79 \\
      \midrule
      Mean & 0.04 & 0.53 & 0.67 & 0.84 \\
      \bottomrule
    \end{tabular}
  \end{adjustbox}
\end{table}

%% file: figures/figure_reflection_separation_float.tex
\begin{figure}[tbp]
  \centering
  \includegraphics[width=\columnwidth]{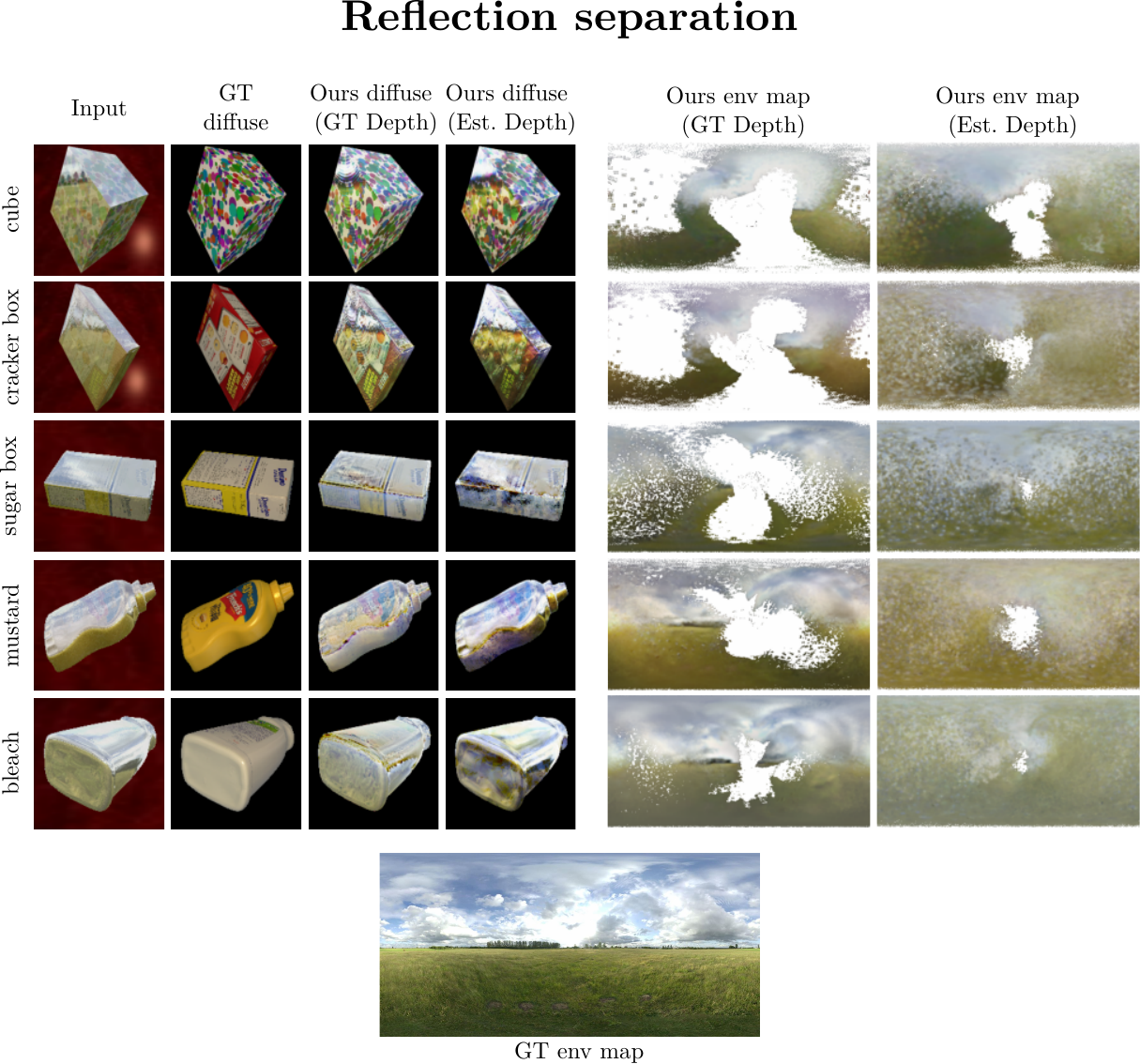}
  \caption[Reflection separation under ground truth and estimated
  depth]{\textbf{Reflection separation} at $r{=}0.7$, frame~20. Each
    object (rows) is decomposed into a diffuse component bound to the
    surface and a reflected component bound to the environment.
    \emph{Left block:} the light field input, the ground truth diffuse
    albedo, and our recovered diffuse under ground truth and estimated
    depth. \emph{Right block:} our recovered environment map under both
    depths, with the ground truth map below. Ground truth depth yields
    the cleaner separation, while the surface normal noise of estimated
    depth blurs the environment map and leaks reflected content into
  the diffuse component.}
  \label{fig:reflection_separation}
\end{figure}

%% file: tables/ablation_combined.tex
\begin{table}[t]
  \caption[Component ablation across reflectivity for objects and
  cube]{\textbf{Component ablation across the full grid}, on both
    geometries and all four reflectivity levels, in the \emph{Sensor}
    regime of \cref{tab:supp_unified}. Within each block the first row
    swaps the ground truth masks for our predicted segmentation,
    \textbf{Ours} is the full pipeline as reported in
    \cref{tab:supp_unified}, each subsequent row removes one further
    component, and the last replaces light field depth with sensor
    depth; the best value per column and block is in bold. The ordering
    is consistent throughout the grid: swapping in sensor depth is the
    most damaging change everywhere, and photometric refinement and
    reflection separation contribute increasingly with reflectivity,
  most sharply on the fully reflective cube.}
  \label{tab:ablation_combined}
  \centering
  \footnotesize
  \setlength{\tabcolsep}{3pt}
  \renewcommand{\arraystretch}{1.05}
  \adjustbox{max width=\columnwidth,center}{%
    \begin{tabular}{c l cccc cccc}
      \toprule
      & & \multicolumn{4}{c}{Objects} &
      \multicolumn{4}{c}{Cube} \\
      \cmidrule(lr){3-6}\cmidrule(lr){7-10}
      $r$ & Variant
      & ADD-S$\uparrow$ & ADD$\uparrow$ &
      ATE\,[m]$\downarrow$ & ARE\,[$^\circ$]$\downarrow$
      & ADD-S$\uparrow$ & ADD$\uparrow$ &
      ATE\,[m]$\downarrow$ & ARE\,[$^\circ$]$\downarrow$ \\
      \midrule
      \multirow{5}{*}{$r=0.0$}
      & Ours w/ predicted masks & 0.902 & 0.818 & \textbf{0.018} &
      10.2 & \textbf{0.849} & \textbf{0.848} & \textbf{0.013} & \textbf{4.2} \\
      & \textbf{Ours} & 0.898 & 0.815 & 0.019 & 10.3 & 0.841 & 0.841
      & 0.014 & 4.4 \\
      & $-$ photometric refinement & 0.901 & 0.821 & \textbf{0.018} &
      9.7 & 0.839 & 0.839 & \textbf{0.013} & 4.9 \\
      & $-$ reflection separation & \textbf{0.903} & \textbf{0.826} &
      0.019 & \textbf{9.1} & 0.837 & 0.837 & \textbf{0.013} & 5.0 \\
      & $-$ light field depth ($\rightarrow$ sensor) & 0.822 & 0.658
      & 0.041 & 19.4 & 0.756 & 0.756 & 0.022 & 7.2 \\
      \midrule
      \multirow{5}{*}{$r=0.5$}
      & Ours w/ predicted masks & 0.885 & 0.784 & 0.019 & 11.8 &
      0.816 & 0.816 & 0.016 & \textbf{4.8} \\
      & \textbf{Ours} & \textbf{0.893} & \textbf{0.799} &
      \textbf{0.017} & \textbf{11.4} & \textbf{0.819} &
      \textbf{0.819} & 0.016 & \textbf{4.8} \\
      & $-$ photometric refinement & 0.881 & 0.752 & 0.021 & 13.5 &
      0.815 & 0.815 & \textbf{0.013} & 5.9 \\
      & $-$ reflection separation & 0.880 & 0.750 & 0.021 & 13.7 &
      0.817 & 0.817 & \textbf{0.013} & 5.8 \\
      & $-$ light field depth ($\rightarrow$ sensor) & 0.827 & 0.589
      & 0.035 & 19.8 & 0.674 & 0.673 & 0.027 & 9.7 \\
      \midrule
      \multirow{5}{*}{$r=0.7$}
      & Ours w/ predicted masks & 0.877 & 0.771 & \textbf{0.021} & 12.1 &
      \textbf{0.783} & \textbf{0.781} & 0.018 & \textbf{6.5} \\
      & \textbf{Ours} & \textbf{0.881} & \textbf{0.774} & \textbf{0.021}
      & \textbf{11.6} & 0.782 & \textbf{0.781} & 0.017 & 6.6 \\
      & $-$ photometric refinement & 0.872 & 0.737 & 0.025 & 12.8 &
      0.768 & 0.768 & \textbf{0.015} & 7.7 \\
      & $-$ reflection separation & 0.875 & 0.739 & 0.025 & 12.6 &
      0.767 & 0.767 & 0.016 & 7.7 \\
      & $-$ light field depth ($\rightarrow$ sensor) & 0.829 & 0.601
      & 0.036 & 20.6 & 0.501 & 0.498 & 0.047 & 17.0 \\
      \midrule
      \multirow{5}{*}{$r=1.0$}
      & Ours w/ predicted masks & 0.878 & 0.769 & 0.023 & 11.6 &
      0.307 & 0.303 & 0.276 & 36.9 \\
      & \textbf{Ours} & \textbf{0.885} & \textbf{0.783} & \textbf{0.022}
      & \textbf{11.0} & \textbf{0.510} & \textbf{0.510} &
      \textbf{0.075} & \textbf{16.8} \\
      & $-$ photometric refinement & 0.856 & 0.691 & 0.031 & 14.3 &
      0.424 & 0.419 & 0.085 & 20.6 \\
      & $-$ reflection separation & 0.854 & 0.685 & 0.031 & 15.3 &
      0.442 & 0.436 & 0.081 & 19.8 \\
      & $-$ light field depth ($\rightarrow$ sensor) & 0.661 & 0.424
      & 0.075 & 30.3 & 0.125 & 0.122 & 0.142 & 64.4 \\
      \bottomrule
  \end{tabular}}
\end{table}

%% file: tables/real_captures_ablation.tex
\begin{table}[!htbp]
  \caption[Component ablation on the captured sequences]{\textbf{Component
    ablation on the captured sequences}. ADD-S as AUC, ATE
    in m, ARE in degrees. Each row below \textbf{Ours} removes one
    component. Replacing light field depth is the most damaging
    change, and reflection separation carries the rotation gain on the
  foil.}
  \label{tab:real_captures_ablation}
  \centering
  \footnotesize
  \setlength{\tabcolsep}{4pt}
  \begin{adjustbox}{max width=\linewidth}
    \begin{tabular}{lcccccc}
      \toprule
      & \multicolumn{3}{c}{mustard bottle}
      & \multicolumn{3}{c}{same bottle, foil-wrapped} \\
      \cmidrule(lr){2-4}\cmidrule(lr){5-7}
      Variant & ADD-S~$\uparrow$ & ATE\,[m]~$\downarrow$
      & ARE\,[$^\circ$]~$\downarrow$
      & ADD-S~$\uparrow$ & ATE\,[m]~$\downarrow$
      & ARE\,[$^\circ$]~$\downarrow$ \\
      \midrule
      \textbf{Ours} & \textbf{0.925} & 0.013 & \textbf{14.0}
      & \textbf{0.941} & \textbf{0.009} & \textbf{6.9} \\
      $-$ photometric refinement & 0.918 & 0.013 & 15.5
      & 0.929 & 0.011 & \textbf{6.9} \\
      $-$ reflection separation & 0.924 & \textbf{0.010} & 18.8
      & 0.904 & 0.017 & 16.2 \\
      $-$ light field depth & 0.780 & 0.039 & 30.1
      & 0.820 & 0.036 & 29.1 \\
      \bottomrule
    \end{tabular}
  \end{adjustbox}
\end{table}

%% file: figures/figure_ablation_float.tex
\begin{figure}[tbp]
  \centering
  \includegraphics[width=\columnwidth]{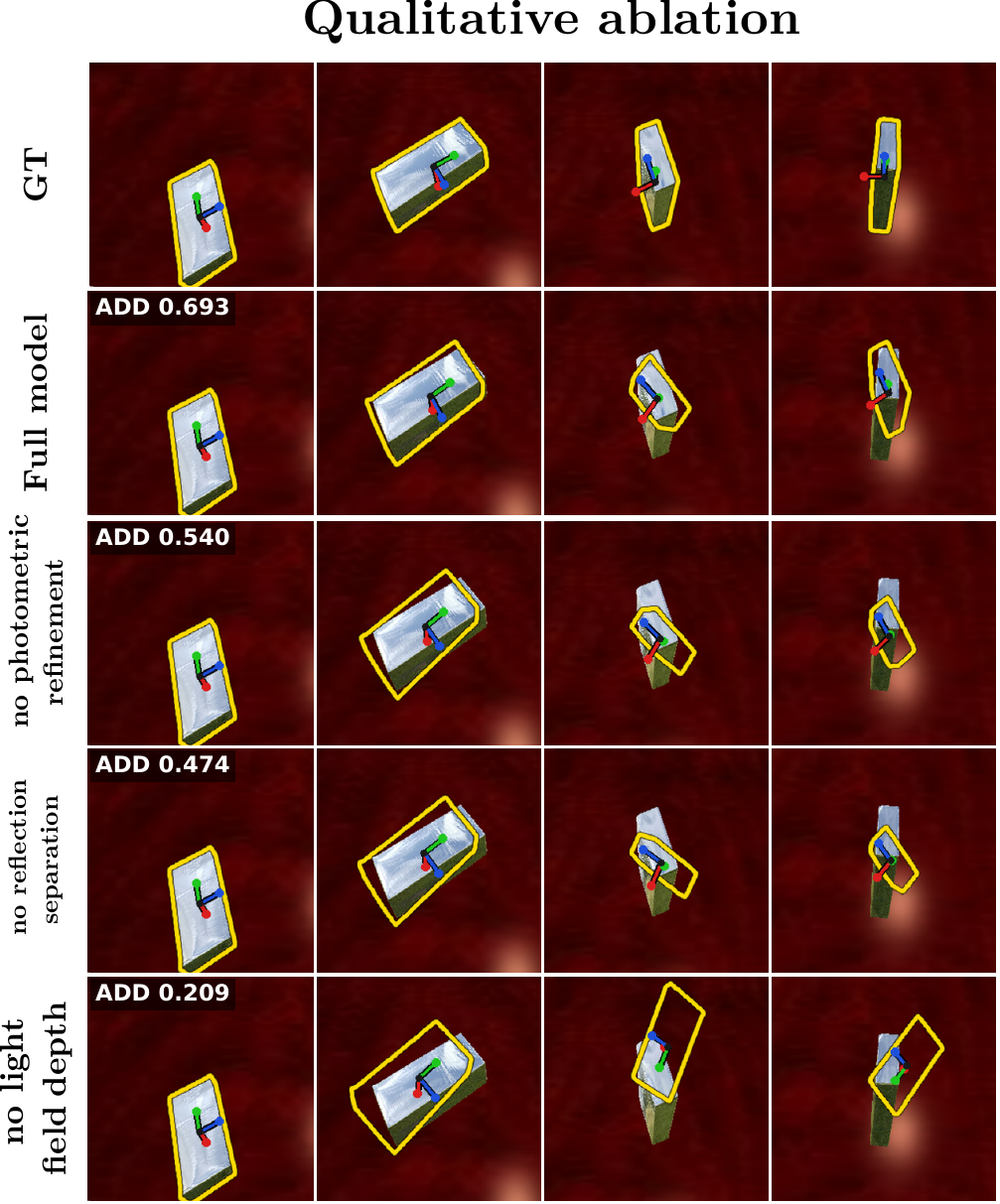}
  \caption[Qualitative pose tracking ablation]{\textbf{Qualitative
    ablation on the most reflective \texttt{sugar\_box} sequence.}
    Each row tracks the object over time (columns left to right), with
    the yellow outline and axes showing the estimated pose and the
    row's mean ADD over this sequence. The top row is the ground truth,
    below it the full model and then variants that each remove one
    component (photometric refinement, reflection separation, and
    finally light field depth replaced by sensor depth). Drift grows as
    each component is removed, most severely once sensor depth replaces
  the light field depth.}
  \label{fig:qualitative_ablation}
\end{figure}

%% file: main.bbl
\begin{thebibliography}{10}
\providecommand{\url}[1]{\texttt{#1}}
\providecommand{\urlprefix}{URL }
\providecommand{\doi}[1]{https://doi.org/#1}

\bibitem{adato2010shape}
Adato, Y., Vasilyev, Y., Zickler, T., Ben-Shahar, O.: {Shape from Specular
  Flow}. IEEE TPAMI  \textbf{32}(11),  2054--2070 (2010)

\bibitem{alperovich2016variational}
Alperovich, A., Goldluecke, B.: {A Variational Model for Intrinsic Light Field
  Decomposition}. In: ACCV. pp. 66--82 (2016)

\bibitem{besl1992method}
Besl, P.J., McKay, N.D.: {A Method for Registration of 3-D Shapes}. IEEE TPAMI
  \textbf{14}(2),  239--256 (1992)

\bibitem{blinn1976texture}
Blinn, J.F., Newell, M.E.: {Texture and Reflection in Computer Generated
  Images}. Commun. ACM  \textbf{19}(10),  542--547 (1976)

\bibitem{canas2009linear}
Ca{\~n}as, G.D., Vasilyev, Y., Adato, Y., Zickler, T., Gortler, S.J.,
  Ben-Shahar, O.: {A Linear Formulation of Shape from Specular Flow}. In: ICCV.
  pp. 191--198 (2009)

\bibitem{chandraker2013object}
Chandraker, M., Reddy, D., Wang, Y., Ramamoorthi, R.: {What Object Motion
  Reveals About Shape With Unknown BRDF and Lighting}. In: CVPR. pp. 2523--2530
  (2013)

\bibitem{OccCasNet_2024}
Chao, W., Duan, F., Wang, X., Wang, Y., Lu, K., Wang, G.: {OccCasNet:
  Occlusion-Aware Cascade Cost Volume for Light Field Depth Estimation}. IEEE
  TCI  \textbf{10},  1680--1691 (2024)

\bibitem{cheng2023putting}
Cheng, H.K., Oh, S.W., Price, B., Lee, J.Y., Schwing, A.: {Putting the Object
  Back into Video Object Segmentation}. In: CVPR (2024)

\bibitem{dansereau2004gradient}
Dansereau, D.G., Bruton, L.T.: {Gradient-based Depth Estimation from 4D Light
  Fields}. In: ISCAS. pp. 549--552 (2004)

\bibitem{dansereau2019liff}
Dansereau, D.G., Girod, B., Wetzstein, G.: {LiFF: Light Field Features in Scale
  and Depth}. In: CVPR. pp. 8042--8051 (2019)

\bibitem{fan2025spectromotion}
Fan, C.D., Chang, C.W., Liu, Y.R., Lee, J.Y., Huang, J.L., Tseng, Y.C., Liu,
  Y.L.: {SpectroMotion: Dynamic 3D Reconstruction of Specular Scenes}. In:
  CVPR. pp. 21328--21338 (2025)

\bibitem{fischler1981random}
Fischler, M.A., Bolles, R.C.: {Random Sample Consensus: A Paradigm for Model
  Fitting with Applications to Image Analysis and Automated Cartography}.
  Commun. ACM  \textbf{24}(6),  381--395 (1981)

\bibitem{gao2024relightable}
Gao, J., Gu, C., Lin, Y., Li, Z., Zhu, H., Cao, X., Zhang, L., Yao, Y.:
  {Relightable 3D Gaussians: Realistic Point Cloud Relighting with BRDF
  Decomposition and Ray Tracing}. In: ECCV (2024)

\bibitem{goncharov2025light}
Goncharov, N., Gray, J.L., Dansereau, D.G.: {Light Field Based 6DoF Tracking of
  Previously Unobserved Objects}. In: MM-SpatialAI Workshop: Multi-Modal
  Spatial AI for Robust Navigation and Open-World Understanding, ICRA (2026)

\bibitem{hirschmuller2008stereo}
Hirschm{\"u}ller, H.: {Stereo Processing by Semiglobal Matching and Mutual
  Information}. IEEE TPAMI  \textbf{30}(2),  328--341 (2008)

\bibitem{kerbl20233d}
Kerbl, B., Kopanas, G., Leimk{\"u}hler, T., Drettakis, G.: {3D Gaussian
  Splatting for Real-Time Radiance Field Rendering}. ACM TOG  \textbf{42}(4),
  139:1--139:14 (2023)

\bibitem{labbe2022megapose}
Labb{\'e}, Y., Manuelli, L., Mousavian, A., Tyree, S., Birchfield, S.,
  Tremblay, J., Carpentier, J., Aubry, M., Fox, D., Sivic, J.: {MegaPose: 6D
  Pose Estimation of Novel Objects via Render \& Compare}. In: CoRL. pp.
  715--725 (2022)

\bibitem{lepetit2009ep}
Lepetit, V., Moreno-Noguer, F., Fua, P.: {EPnP: An Accurate O(n) Solution to
  the PnP Problem}. IJCV  \textbf{81}(2),  155--166 (2009)

\bibitem{li2018robust}
Li, T., Lun, D.P.K., Chan, Y.H., Budianto, B.: {Robust Reflection Removal Based
  on Light Field Imaging}. IEEE TIP  \textbf{28}(4),  1798--1812 (2019)

\bibitem{lin2024sam}
Lin, J., Liu, L., Lu, D., Jia, K.: {SAM-6D: Segment Anything Model Meets
  Zero-Shot 6D Object Pose Estimation}. In: CVPR. pp. 27906--27916 (2024)

\bibitem{liu2024grounding}
Liu, S., Zeng, Z., Ren, T., Li, F., Zhang, H., Yang, J., Jiang, Q., Li, C.,
  Yang, J., Su, H., Zhu, J., Zhang, L.: {Grounding DINO: Marrying DINO with
  Grounded Pre-Training for Open-Set Object Detection}. In: ECCV. pp. 38--55
  (2024)

\bibitem{liu2024bigs}
Liu, Z., Guo, Y., Li, X., Bickel, B., Zhang, R.: {BiGS: Bidirectional
  Primitives for Relightable 3D Gaussian Splatting}. In: 3DV (2025)

\bibitem{lucas1981iterative}
Lucas, B.D., Kanade, T.: {An Iterative Image Registration Technique with an
  Application to Stereo Vision}. In: IJCAI. pp. 674--679 (1981)

\bibitem{mildenhall2021nerf}
Mildenhall, B., Srinivasan, P.P., Tancik, M., Barron, J.T., Ramamoorthi, R.,
  Ng, R.: {NeRF: Representing Scenes as Neural Radiance Fields for View
  Synthesis}. In: ECCV. pp. 405--421 (2020)

\bibitem{naylor2026surf}
Naylor, J., Ila, V., Dansereau, D.G.: {Surf-NeRF: Surface regularised neural
  radiance fields}. In: Proceedings of the IEEE/CVF Conference on Computer
  Vision and Pattern Recognition. pp. 258--268 (2026)

\bibitem{park2017coloredicp}
Park, J., Zhou, Q.Y., Koltun, V.: {Colored Point Cloud Registration Revisited}.
  In: ICCV. pp. 143--152 (2017)

\bibitem{park2021hypernerf}
Park, K., Sinha, U., Hedman, P., Barron, J.T., Bouaziz, S., Goldman, D.B.,
  Martin-Brualla, R., Seitz, S.M.: {HyperNeRF: A Higher-Dimensional
  Representation for Topologically Varying Neural Radiance Fields}. ACM TOG
  \textbf{40}(6),  238:1--238:12 (2021)

\bibitem{phong1998illumination}
Phong, B.T.: {Illumination for Computer Generated Pictures}. Commun. ACM
  \textbf{18}(6),  311--317 (1975)

\bibitem{pumarola2021d}
Pumarola, A., Corona, E., Pons-Moll, G., Moreno-Noguer, F.: {D-NeRF: Neural
  Radiance Fields for Dynamic Scenes}. In: CVPR. pp. 10318--10327 (2021)

\bibitem{ravi2025sam}
Ravi, N., Gabeur, V., Hu, Y.T., Hu, R., Ryali, C., Ma, T., Khedr, H.,
  R{\"a}dle, R., Rolland, C., Gustafson, L., Mintun, E., Pan, J., Alwala, K.V.,
  Carion, N., Wu, C.Y., Girshick, R., Doll{\'a}r, P., Feichtenhofer, C.: {SAM
  2: Segment Anything in Images and Videos}. In: ICLR (2025)

\bibitem{ravi2020accelerating}
Ravi, N., Reizenstein, J., Novotny, D., Gordon, T., Lo, W.Y., Johnson, J.,
  Gkioxari, G.: {Accelerating 3D Deep Learning with PyTorch3D} (2020)

\bibitem{roth2006specular}
Roth, S., Black, M.J.: {Specular Flow and the Recovery of Surface Structure}.
  In: CVPR. pp. 1869--1876 (2006)

\bibitem{sankaranarayanan2010specular}
Sankaranarayanan, A.C., Veeraraghavan, A., Tuzel, O., Agrawal, A.: {Specular
  Surface Reconstruction from Sparse Reflection Correspondences}. In: CVPR. pp.
  599--606 (2010)

\bibitem{shafer1985using}
Shafer, S.A.: {Using Color to Separate Reflection Components}. Color Research
  \& Application  \textbf{10}(4),  210--218 (1985)

\bibitem{shen2023light}
Shen, Z., Zhang, S., Lin, Y.: {Light Field Reflection and Background Separation
  Network Based on Adaptive Focus Selection}. IEEE TCI  \textbf{9},  435--447
  (2023)

\bibitem{srinivasan2021nerv}
Srinivasan, P.P., Deng, B., Zhang, X., Tancik, M., Mildenhall, B., Barron,
  J.T.: {NeRV: Neural Reflectance and Visibility Fields for Relighting and View
  Synthesis}. In: CVPR. pp. 7495--7504 (2021)

\bibitem{sulc2016reflection}
Sulc, A., Alperovich, A., Marniok, N., Goldluecke, B.: {Reflection Separation
  in Light Fields Based on Sparse Coding and Specular Flow}. In: VMV. pp.
  137--144 (2016)

\bibitem{sun2021loftr}
Sun, J., Shen, Z., Wang, Y., Bao, H., Zhou, X.: {LoFTR: Detector-Free Local
  Feature Matching with Transformers}. In: CVPR. pp. 8922--8931 (2021)

\bibitem{tao2013depth}
Tao, M.W., Hadap, S., Malik, J., Ramamoorthi, R.: {Depth from Combining Defocus
  and Correspondence Using Light-Field Cameras}. In: ICCV. pp. 673--680 (2013)

\bibitem{tao2015depth}
Tao, M.W., Su, J.C., Wang, T.C., Malik, J., Ramamoorthi, R.: {Depth Estimation
  and Specular Removal for Glossy Surfaces Using Point and Line Consistency
  with Light-Field Cameras}. IEEE TPAMI  \textbf{38}(6),  1155--1169 (2016)

\bibitem{wang2019normalized}
Wang, H., Sridhar, S., Huang, J., Valentin, J., Song, S., Guibas, L.J.:
  {Normalized Object Coordinate Space for Category-Level 6D Object Pose and
  Size Estimation}. In: CVPR. pp. 2642--2651 (2019)

\bibitem{wang2015occlusion}
Wang, T.C., Efros, A.A., Ramamoorthi, R.: {Occlusion-Aware Depth Estimation
  Using Light-Field Cameras}. In: ICCV. pp. 3487--3495 (2015)

\bibitem{wen2020se}
Wen, B., Mitash, C., Ren, B., Bekris, K.E.: {se(3)-TrackNet: Data-driven 6D
  Pose Tracking by Calibrating Image Residuals in Synthetic Domains}. In: IROS.
  pp. 10367--10373 (2020)

\bibitem{wen2023bundlesdf}
Wen, B., Tremblay, J., Blukis, V., Tyree, S., M{\"u}ller, T., Evans, A., Fox,
  D., Kautz, J., Birchfield, S.: {BundleSDF: Neural 6-DoF Tracking and 3D
  Reconstruction of Unknown Objects}. In: CVPR. pp. 606--617 (2023)

\bibitem{wen2024foundationpose}
Wen, B., Yang, W., Kautz, J., Birchfield, S.: {FoundationPose: Unified 6D Pose
  Estimation and Tracking of Novel Objects}. In: CVPR. pp. 17868--17879 (2024)

\bibitem{wu20244d}
Wu, G., Yi, T., Fang, J., Xie, L., Zhang, X., Wei, W., Liu, W., Tian, Q., Wang,
  X.: {4D Gaussian Splatting for Real-Time Dynamic Scene Rendering}. In: CVPR.
  pp. 20310--20320 (2024)

\bibitem{xiang_posecnn_2017}
Xiang, Y., Schmidt, T., Narayanan, V., Fox, D.: {PoseCNN: A Convolutional
  Neural Network for 6D Object Pose Estimation in Cluttered Scenes}. In: RSS
  (2018)

\end{thebibliography}
